\documentclass[10pt,twocolumn,letterpaper]{article}

\usepackage{cvpr}              

\graphicspath{{figures/}}

\usepackage{color, colortbl}
\definecolor{ColumnColor}{rgb}{0.93, 0.93,0.98}
\definecolor{ColumnColor2}{rgb}{1.0, 0.95,0.95}

\usepackage{amsthm}

\newtheorem{remark}{Remark}
\newtheorem{theorem}{Theorem}

\newtheorem{proposition}{Proposition}

\newcommand{\bfr}{\boldsymbol{r}}

\newcommand{\bfx}{\boldsymbol{x}}
\newcommand{\bfy}{\boldsymbol{y}}

\newcommand{\bfM}{\boldsymbol{M}}

\newcommand{\bftheta}{\boldsymbol{\theta}}

\newif\ifDummy

\definecolor{cvprblue}{rgb}{0.21,0.49,0.74}
\usepackage[pagebackref,breaklinks,colorlinks,allcolors=cvprblue]{hyperref}

\def\paperID{*****} 
\def\confName{CVPR}
\def\confYear{2026}

\title{Noise \& Pattern: Identity-Anchored Tikhonov Regularization\\ for Robust Structural Anomaly Detection}

\author{
Alexander Bauer$^{1,2}$\\
$^{1}$Machine Learning Group, TU Berlin\\
$^{2}$BIFOLD, Berlin, Germany\\
{\tt\small alexander.bauer@tu-berlin.de}
\and
Klaus-Robert Müller$^{1,2,3,4}$\\
$^{3}$Dept.\ of AI, Korea University, Seoul, Republic of Korea\\
$^{4}$MPI for Informatics, Saarbrücken, Germany\\
{\tt\small klaus-robert.mueller@tu-berlin.de}
}

\begin{document}
\maketitle
\begin{abstract}
Anomaly detection plays a pivotal role in automated industrial inspection, aiming to identify subtle or rare defects in otherwise uniform visual patterns. As collecting representative examples of all possible anomalies is infeasible, we tackle structural anomaly detection using a self-supervised autoencoder that learns to repair corrupted inputs. To this end, we introduce a corruption model that injects artificial disruptions into training images to mimic structural defects. While reminiscent of denoising autoencoders, our approach differs in two key aspects. First, instead of unstructured i.i.d.\ noise, we apply structured, spatially coherent perturbations that make the task a hybrid of segmentation and inpainting. Second, and counterintuitively, we add and preserve Gaussian noise on top of the occlusions, which acts as a Tikhonov regularizer anchoring the Jacobian of the reconstruction function toward identity. This identity-anchored regularization stabilizes reconstruction and further improves both detection and segmentation accuracy. On the MVTec AD benchmark, our method achieves state-of-the-art results (I/P-AUROC: 99.9/99.4), supporting our theoretical framework and demonstrating its practical relevance for automatic inspection.
\end{abstract}
\section{Introduction}
\label{sec:intro}
Anomaly Detection (AD) aims to identify patterns that deviate from a defined notion of normality---a task of considerable importance in domains such as industrial quality control, medical imaging, and beyond~\cite{HaselmannGT18, BergmannLFSS19, LSR, BergmannFSS20, VenkataramananP20, SchleglSWSL17, NapoletanoPS18, texture2016, LiuLZKWBRC20, RothPZSBG22, WanGLW22, abs-1806-04972, SchleglSWLS19, Kohlbrenner0NBS20, HonerNBMG17, HonerNBMG17, TanHDSRK21, ZimmererIPKM19, AbatiPCC19, RuffKVMSKDM21, DefardSLA20}. 
A wide range of approaches has been explored for AD. 
Early methods rely on classical unsupervised techniques such as Principal Component Analysis (PCA)~\cite{Hotelling33, ScholkopfSM98, Hoffmann07}, One-Class Support Vector Machines (OC-SVM)~\cite{ScholkopfPSSW01}, Support Vector Data Description (SVDD)~\cite{TaxD04}, nearest-neighbor algorithms~\cite{KnorrNT00, RamaswamyRS00}, and Kernel Density Estimation (KDE)~\cite{Parzen62}. 
In recent years, research has shifted toward deep learning-based models, including autoencoders~\cite{PrincipiVSP17, abs-1806-04972, ChalapathyMC17, KieuYGJ19, ZhouP17, ZongSMCLCC18, KimSLJCKY20, DengZMS13, VenkataramananP20}, deep one-class classifiers~\cite{ErfaniRKL16, KimKYC23, RuffGDSVBMK18}, generative approaches~\cite{SchleglSWSL17, SchleglSWLS19}, and self-supervised methods~\cite{GolanE18, TackMJS20, LSR, HaselmannGT18, ZavrtanikKS212, ZavrtanikKS21, LiSYP21, InTra}.
\begin{figure}[b]
\centering
\includegraphics[scale = 0.44]{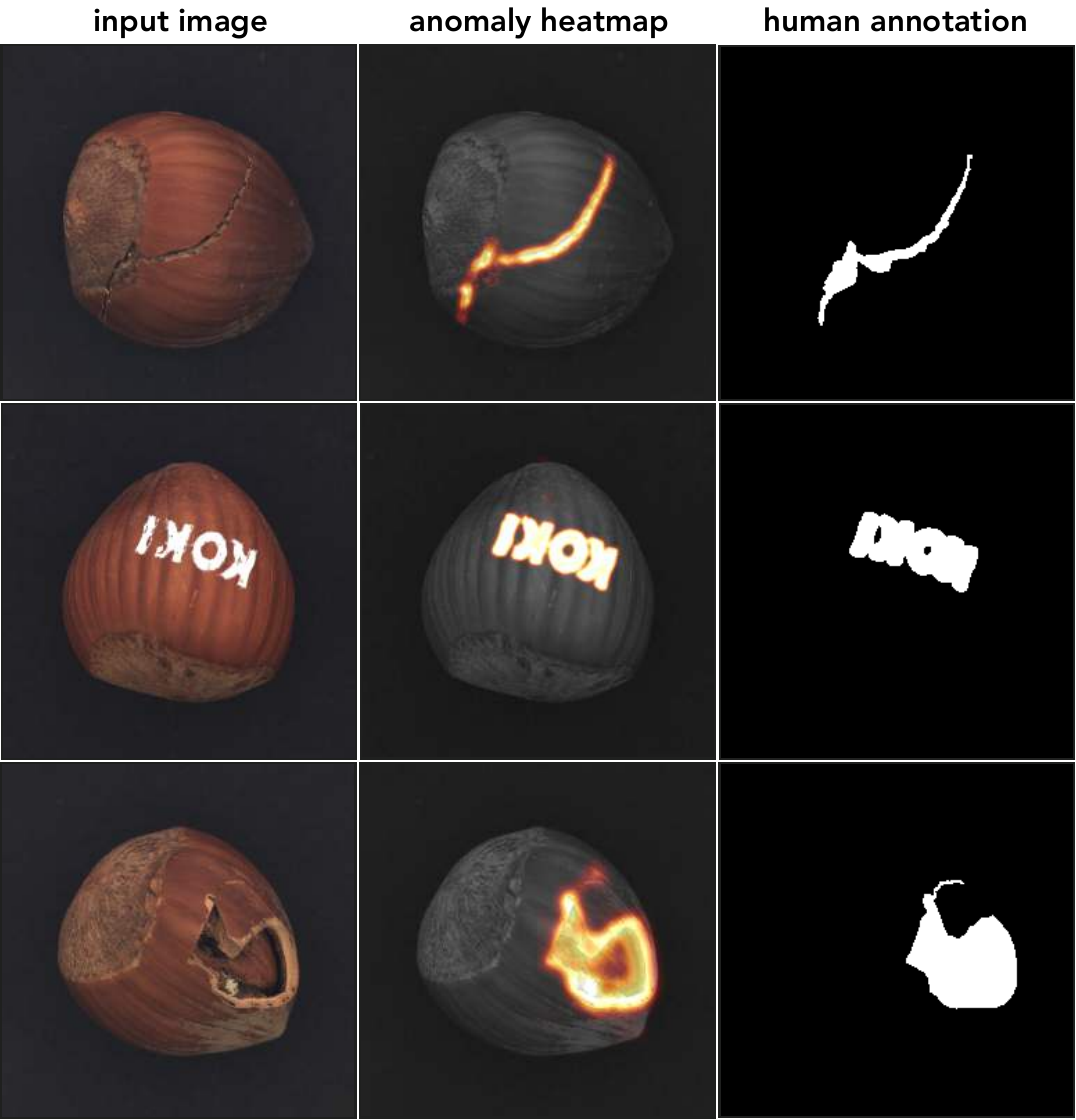}
\caption{
\textbf{Robust Structural Anomaly Detection in a Nutshell.}
Our qualitative results on MVTec~AD (\emph{hazelnut}): input images, predicted heatmaps, and ground-truth masks.
}
\label{fig_nuts}
\end{figure}

In this work, we focus on AD in static images, which is typically addressed at two levels: 
(a) \emph{image-level} detection, formulated as a binary classification problem, and 
(b) \emph{pixel-level} detection, formulated as a segmentation task to localize anomalous regions. 
Specifically, we consider \emph{structural anomalies}, which manifest as localized disruptions in texture or form 
(e.g., scratches, deformations, or material defects). 
Due to their spatially local nature at the pixel level, structural anomalies are well-suited for reconstruction-based methods. 
In contrast, \emph{logical anomalies} refer to semantic inconsistencies at the scene level 
(e.g., incorrect object counts or implausible spatial relations), which are global and not uniquely localizable, 
and thus fall outside the scope of reconstruction-based approaches \cite{BergmannBFSS22}.

A prominent line of AD research builds upon neural image completion 
\cite{PathakKDDE16, IizukaS017, Yu0YSLH18, LiuRSWTC18, YuLYSLH19, abs-1802-05798}, 
where models are trained in a self-supervised manner to reconstruct missing or altered image regions. 
Representative examples include LSR~\cite{LSR}, RIAD~\cite{ZavrtanikKS21}, CutPaste~\cite{LiSYP21}, 
InTra~\cite{InTra}, DRAEM~\cite{ZavrtanikKS212}, and SimpleNet~\cite{LiuZXW23}. 
Despite architectural differences, these approaches share the common objective of restoring normal content from corrupted inputs. 

Our work follows this reconstruction paradigm but extends it with a more expressive \emph{corruption model}. 
Previous approaches simulated anomalies ranging from simple patch removal~\cite{ZavrtanikKS21} 
to Perlin noise-based texture replacements~\cite{ZavrtanikKS212}, 
yet offered only limited structural diversity. 
In contrast, our model spans multiple dimensions of variation---\emph{texture}, \emph{geometric shape}, and \emph{opacity}---to generate 
a broader range of structural perturbations. 
This diversity (paired with noise regularization described below) is crucial for generalization to real-world anomalies.

We train an autoencoder to repair such artificially corrupted images by minimizing the reconstruction loss between clean and corrupted counterparts. 
A key element of our approach is the contrasting interplay between persistent structural deviations and Gaussian noise applied across the entire image. 
This combination not only regularizes training but also enhances robustness in anomaly localization. 
It further improves detection accuracy by increasing reconstruction fidelity in both corrupted and normal regions, enabling the model to disentangle the true structure of normal patterns from unstructured pixel-wise variations. 

This noise component also admits a clear theoretical interpretation: it acts as a Tikhonov-like regularization that anchors the Jacobian of the reconstruction function toward the identity. 
As a result, we obtain a \emph{Filtering Autoencoder} (FAE) that removes anomalous patterns while preserving normal variability, thereby enabling robust and accurate anomaly detection. 
Figure~\ref{fig_nuts} provides a visual illustration of the detection performance of our method.

The remainder of the paper is organized as follows.
Section~\ref{sec:section2} introduces the training framework, 
Section~\ref{sec:section3} provides the theoretical justification linking our approach to identity-anchored Tikhonov regularization, 
Section~\ref{sec:section4} details the corruption model, 
and Section~\ref{sec:section5} presents experimental results. 
We conclude in Section~\ref{sec:section6}.

\section{Methodology}
\label{sec:section2}
We model the autoencoder as a parameterized function
\[
f_{\bftheta} : [0,1]^{h \times w \times 3} \rightarrow [0,1]^{h \times w \times 3},
\]
with learnable parameters $\bftheta$ and input-output tensors corresponding to RGB images of spatial size $h \times w$.  
Let $\bfx \in [0,1]^{h \times w \times 3}$ denote an anomaly-free training image.  
A corrupted variant $\hat{\bfx}$ is created by modifying selected regions, which are indicated by a real-valued mask $\bfM \in [0,1]^{h \times w \times 3}$.  
Its complement is written as $\bar{\bfM} := \mathbf{1} - \bfM$, where $\mathbf{1}$ is the all-ones tensor.
Note that $\bfM$ is not binary and takes on values from a continuous range.\\

\noindent \textbf{Training.}  We start with a simpler objective that will later be modified. 
Following~\cite{math12243988}, we minimize a loss $\mathcal{L}(\hat{\bfx}, \bfx, \bfM; \bftheta)$ over the model weights $\bftheta$ defined as
\begin{equation*}
\label{E_12041904}
\frac{1-\lambda}{\|\bar{\bfM}\|_1}\|\bar{\bfM}\odot(f_{\bftheta}(\hat{\bfx})-\bfx)\|_2^2 
+ \frac{\lambda}{\|\bfM\|_1}\|\bfM\odot(f_{\bftheta}(\hat{\bfx})-\bfx)\|_2^2,
\end{equation*}
where $\odot$ denotes elementwise multiplication, $\|\cdot\|_p$ the standard $\ell^p$-norm on tensors, and $\lambda\!\in\![0,1]$ is a hyperparameter.

The above objective is divided into two terms to penalize prediction errors on corrupted and uncorrupted regions differently.
For simplicity, we omit these weights in the following discussion and adopt the simplified form:
\begin{equation}
\label{E_17091954}
\mathcal{L}_{\text{DAE}}(\hat{\bfx}, \bfx; \bftheta) =  \|f_{\bftheta}(\hat{\bfx}) - \bfx\|_2^2,
\end{equation}
which resembles the denoising autoencoder (DAE) \cite{VincentLBM08}.\\

\noindent \textbf{Test-time detection.} Once the autoencoder $f_{\bftheta}$ is trained, anomaly maps can be derived from the discrepancy between an input $\hat{\bfx}$ and its reconstruction $f_{\bftheta}(\hat{\bfx})$.  
Following~\cite{math12243988}, we define this pixel-based discrepancy as a mapping
\[
\Delta : [0,1]^{h \times w \times 3} \times [0,1]^{h \times w \times 3} \rightarrow [0,1]^{h \times w}.
\]
There are multiple options for $\Delta$
including pixel-wise Mean Squared Error (MSE), Structural Similarity Index Measure (SSIM) \cite{BergmannLFSS19} and Gradient Magnitude Similarity (GMS) \cite{XueZMB14}.
Thresholding a corresponding output yields a binary segmentation of anomalies.  

To enhance robustness, we smooth the difference map $\Delta(\cdot, \cdot)$ before thresholding. Specifically, we define
\begin{equation}
\label{eq_anomap}
\text{anomap}_{f_{\bftheta}}^{n,k}(\hat{\bfx}) := G_k^n\!\bigl(\Delta(\hat{\bfx}, f_{\bftheta}(\hat{\bfx}))\bigr),
\end{equation}
where $G_k$ is a mean filter of size $k\times k$ with entries $1/k^2$, and $G_k^n$ denotes $n$ repeated applications (with $G_k^0$ being the identity).  
Hyperparameters $n,k \in \mathbb{N}$ control the amount of smoothing.  
Applying a threshold to $\text{anomap}_{f_{\bftheta}}^{n,k}(\hat{\bfx})$ then produces the anomaly segmentation mask.  

\begin{figure}[t]
\centering
\includegraphics[scale = 0.6]{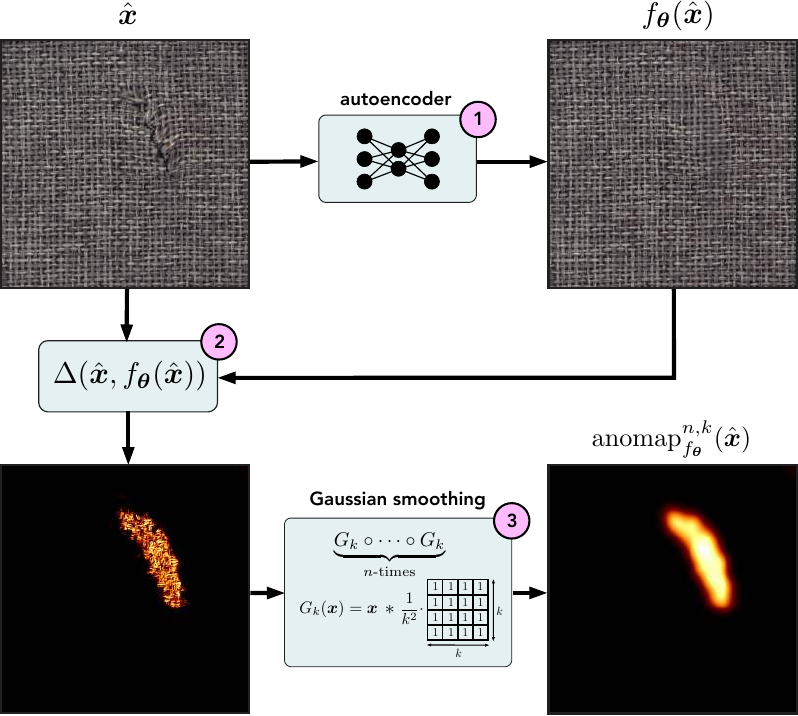}
\caption{
Illustration of our AD process.
Given input $\hat{\bfx}$, our model produces an output $f_{\bftheta}(\hat{\bfx})$,
which replicates normal regions and replaces irregularities with locally consistent patterns.
Then we compute a spatial difference map $\Delta(\hat{\bfx}, f_{\bftheta}(\hat{\bfx})) \in \mathbb{R}^{h \times w}$.
In the last step we apply a series of averaging convolutions $G_k$ to the difference map to produce final anomaly heatmap $\text{anomap}_{f_{\bftheta}}^{n,k}(\hat{\bfx})$.
}
\label{fig_bauer3}
\end{figure}

An image-level anomaly score is obtained by aggregating pixel values of $\text{anomap}_{f_{\bftheta}}^{n,k}(\hat{\bfx})$, typically via summation, though alternative reductions such as the maximum can be used to reduce sensitivity to anomaly size. The full detection workflow (based on \cite{math12243988}) is illustrated in Figure~\ref{fig_bauer3}.\\

\noindent \textbf{Filtering Autoencoder.}
To this end, we propose a new type of regularized autoencoder by modifying the objective in~(\ref{E_17091954}) 
to include isotropic Gaussian noise $\boldsymbol{\epsilon}\!\sim\!\mathcal{N}(\mathbf{0},\sigma^2 I)$:
\begin{equation}
\label{E_17091127}
\mathcal{L}_{\text{FAE}}(\hat{\bfx}, \bfx, \boldsymbol{\epsilon}; \bftheta) 
= \|f_{\bftheta}(\hat{\bfx} + \boldsymbol{\epsilon}) - (\bfx + \boldsymbol{\epsilon})\|_2^2,
\end{equation}
where $\bfx$ is a normal sample, $\hat{\bfx}$ its corrupted version, and $\boldsymbol{\epsilon}$ Gaussian noise.
That is, for each pair $(\hat{\bfx}, \bfx)$, we add small-variance noise to both input and target, 
encouraging the model output $f_{\bftheta}(\hat{\bfx} + \boldsymbol{\epsilon})$ to match the perturbed normal sample $\bfx + \boldsymbol{\epsilon}$.
This seemingly minor modification introduces an effective regularization 
otherwise absent from~(\ref{E_17091954}), as discussed in detail in the next section.

We refer to the model trained with~(\ref{E_17091127}) as the \emph{Filtering Autoencoder} (FAE).
The name reflects its inductive bias: the model preserves normal regions (identity on the manifold) while filtering out structural anomalies. 
Theoretically, this behavior can be viewed as a form of conservative projection~\cite{math12243988} onto the data manifold, 
but we use the term \emph{filtering} to emphasize its practical role in anomaly detection.

\section{Identity-Anchored Tikhonov Regularization}
\label{sec:section3}
Self-supervised autoencoders detect structural anomalies by learning to restore artificially corrupted regions
while leaving normal regions unchanged.
We found that detection performance improves further when the model is additionally trained to preserve small-variance Gaussian noise applied across the entire image after introducing such corruptions.

At first, this may appear counterintuitive, since our strategy resembles a DAE, which is trained to remove rather than preserve noise. However, exposing the model simultaneously to structured corruptions and unstructured Gaussian noise in a complementary manner encourages it to disentangle these effects, thereby enhancing its ability to recognize genuine structural deviations.

A related idea is well established in supervised learning.
For image classification, 
for example, augmenting the data with Gaussian noise 
$\boldsymbol{\epsilon} \sim \mathcal{N}(\mathbf{0}, \sigma^2 I_d)$ 
leads to the training objective
\[
\mathbb{E}_{\boldsymbol{\epsilon}}
  \bigl[ \| f(\bfx + \boldsymbol{\epsilon}) - \bfy \|^2 \bigr].
\]
Bishop~\cite{Bishop95} showed that, up to second order in $\sigma^2$, this objective 
is equivalent to augmenting the clean objective with an additional regularization term according to
\begin{equation}
\label{E_02102100}
\| f(\bfx) - \bfy \|^2 + \sigma^2 \| D_{\bfx} f \|_F^2,
\end{equation}
which is the familiar Tikhonov penalty form.
This penalty enforces local insensitivity to small input perturbations, stabilizing the mapping and improving generalization.

In our case, Gaussian noise serves a similar stabilizing role, but with a crucial and \emph{novel} modification: 
it is added symmetrically to both the input and the target during training. 
This formulation distinguishes the FAE from conventional denoising or contractive autoencoder variants 
and yields a clear theoretical interpretation, which we formalize in the following theorem. 
With a slight abuse of notation, we write $\mathcal{O}(\sigma^4)$ to denote the remainder 
$R \in \mathcal{O}(\sigma^4)$ of the Taylor expansion.
\begin{theorem}[Identity-Anchored Tikhonov Regularization]
\label{thm:ia_tikhonov_pointwise_general}
Let $(\hat{\bfx},\bfx)\sim \mathbb{P}_{\hat{\bfx},\bfx}$ and
$\boldsymbol{\epsilon}\sim\mathcal N(\mathbf 0,\sigma^2 I_d)$ be independent.
Assume $f_{\bftheta} \in C^3$ in a neighborhood of $\hat{\bfx}$ almost surely, and $D^3 f_{\bftheta}$ is locally Lipschitz.
Let $\bfr := f_{\bftheta}(\hat{\bfx}) - \bfx$, 
$J_{\bftheta}(\hat{\bfx}) := D_{\hat{\bfx}} f_{\bftheta}$ denote the Jacobian, 
and $\Delta f_{\bftheta}$ the component-wise Laplacian of $f_{\bftheta}$.
Define
\begin{equation}
\label{E_25091119}
\mathcal{L}(\sigma;\bftheta)
:= \mathbb{E}_{(\hat{\bfx},\bfx),\,\boldsymbol{\epsilon}}
\!\Bigl[\|f_{\bftheta}(\hat{\bfx}+\boldsymbol{\epsilon})-(\bfx+\boldsymbol{\epsilon})\|^2_2\Bigr].
\end{equation}
Then, as $\sigma \to 0$, the total loss $\mathcal{L}(\sigma; \bftheta)$ admits the asymptotic expansion
\begin{equation}
\label{eq:distributional_expansion_final}
\mathbb{E}\!\bigl[ \|\bfr\|^2_2 
+ \sigma^2\big( \|J_{\bftheta}(\hat{\bfx})-I_d\|_F^2 
+ \bfr^\top \Delta f_{\bftheta}(\hat{\bfx}) \big)\bigr]
+\mathcal{O}(\sigma^4),
\end{equation}
provided that $\mathbb{E}\|\bfr\|_2^2, \mathbb{E}\|J_{\bftheta}(\hat{\bfx})\|_F^2, \mathbb{E}\|D^2_{\hat{\bfx}} f_{\bftheta}\|_F^2 < \infty$.
Here, the expectation $\mathbb{E}$ is taken with respect to $(\hat{\bfx}, \bfx)$.
\end{theorem}
Consider the curvature term $\bfr^\top \Delta f$ in~(\ref{eq:distributional_expansion_final}). 
First note that for locally affine mappings, this term vanishes since the Hessian of $f$ is zero. 
This includes, for instance, piecewise-linear architectures such as autoencoders 
with ReLU activations, provided the evaluation point lies within a single linear region. 
In such cases, the expansion simplifies to
\begin{equation}
\label{E_24092037}
\mathbb{E}_{\hat{\bfx}, \bfx}\!\left[
  \|f_{\bftheta}(\hat{\bfx}) - \bfx\|_2^2  
  + \sigma^2 \|J_{\bftheta}(\hat{\bfx}) - I_d\|_F^2
\right]
+ \mathcal{O}(\sigma^4).
\end{equation}

When a nonlinear output activation is used (e.g., a sigmoid in the final layer), the Hessian no longer vanishes, leaving the term $\bfr^\top \Delta f$. 
However, it can be uniformly bounded by a constant multiple of $\|\bfr\| \cdot \|J(\hat{\bfx})\|_F^2$, where the constant depends only on the activation. 
It is therefore proportional to the residual norm $\|\bfr\|$ and becomes negligible as the training error approaches zero. 
Under these favourable conditions, Theorem~\ref{thm:ia_tikhonov_pointwise_general} implies that the loss in~(\ref{E_25091119}) admits an asymptotic expansion approximating~(\ref{E_24092037}). 
As the remainder $\mathcal{O}(\sigma^4)$ vanishes rapidly for $\sigma \to 0$, the dominant regularization effect reduces to the Frobenius term $\|J_{\bftheta}(\hat{\bfx}) - I_d\|_F^2$, 
supporting the interpretation of additive Gaussian noise training as a Tikhonov regularizer that promotes local identity preservation.

\subsection{Complementary Effects of the Jacobian Penalty}
\begin{figure}[t]
\centering
\includegraphics[scale = 0.44]{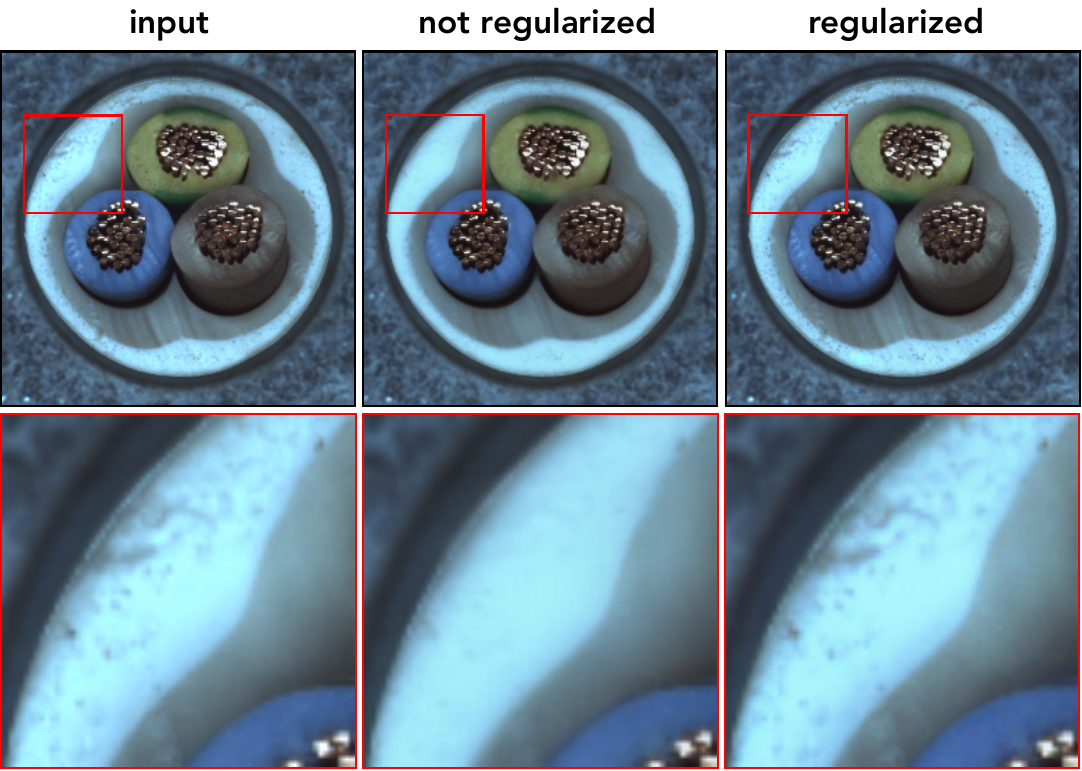}
\caption{
Comparison of reconstruction quality between our models trained with and without Gaussian noise. 
The first row shows an input image $\bfx$ and its reconstructions $f(\bfx)$ from both models. 
The second row provides zoom-ins of the region marked by a red square. 
Reconstructions in normal regions are visibly \textbf{improved} when the model is \textbf{regularized} with Gaussian noise.
}
\label{fig_normal}
\end{figure}

It is instructive to contrast the objective in $(\ref{E_24092037})$ with Bishop’s original result in (\ref{E_02102100}), where
training with noisy inputs and clean targets yields the regularizer
$\|J_{\bftheta}(\bfx)\|_F^2$. This penalty pulls the Jacobian toward zero,
promoting local flatness and insensitivity to perturbations---an
appropriate inductive bias for classification. In our reconstruction
setting, however, such a flatness prior is in tension with the need to
faithfully reproduce inputs along the data manifold.
Here, the identity-anchored penalty $\|J_{\bftheta}-I_d\|_F^2$ adds several complementary effects to the objective summarized below.\\[1pt]
\textbf{Reconstruction of normal regions.}  
A standard autoencoder trained only with reconstruction loss can 
potentially memorize training patterns, yielding spurious mappings that merely interpolate between training examples. 
The regularizer $\|J_{\bftheta}(\hat{\bfx}) - I_d\|_F^2$ counteracts this by normalizing the gradients of $f$ in tangent directions of the data manifold.  
This reduces local collapse (vanishing gradients) or overstretching (exploding gradients) and thereby stabilizes reconstructions of unseen normal data.  
Moreover, if two $C^1$-mappings agree on a manifold, their Jacobians must also agree in tangent directions.
This penalty encourages this necessary local condition for 
$f$ to behave as the identity on the manifold.
As a result, the reconstruction of fine-grained details in normal regions becomes more accurate, as illustrated in Figure~\ref{fig_normal}.\\[1pt]
\textbf{Idempotency.}
Geometrically, our autoencoder is designed to project corrupted inputs 
back onto the manifold of normal samples 
$\mathcal{M} \subset \mathbb{R}^d$ while acting as the identity on $\mathcal{M}$.
Together, these requirements imply that $f$ should be idempotent, i.e.\ $f(f(\hat{\bfx})) = f(\hat{\bfx})$.
To analyze how the regularizer promotes idempotency, we expand $f$ at $\bfx \in \mathcal{M}$ and evaluate at $f(\hat{\bfx})$
where $\hat{\bfx}$ is a corrupted version of $\bfx$.
Since the penalty terms pulls the Jacobian toward identity, assuming the extreme case
$J_{\bftheta}(\bfx) = I_d$ yields
\begin{equation*}
\begin{aligned}
f(f(\hat{\bfx})) &= f(\bfx) + J_{\bftheta}(\bfx)\bigl(f(\hat{\bfx}) - \bfx\bigr) + R_2 \\
&= f(\hat{\bfx}) + (f(\bfx) - \bfx) + R_2,
\end{aligned}
\end{equation*}
where $R_2$ is the second-order Taylor remainder.  
If the Jacobian is $L$-Lipschitz along the segment from $\bfx$ to $f(\hat{\bfx})$, then
$\|R_2\| \le L\|f(\hat{\bfx})-\bfx\|^2$.
Hence,
\begin{equation}
\label{E_22092221}
f(f(\hat{\bfx})) = f(\hat{\bfx}) + \underbrace{(f(\bfx) - \bfx)}_{\text{bias on } \mathcal{M}} + \mathcal{O}\!\big(\underbrace{\|f(\hat{\bfx})-\bfx\|^2}_{\text{reconstruction error}}\big).
\end{equation}
That is, equation~(\ref{E_22092221}) directly links idempotency to the reconstruction accuracy 
and to how well $f$ approximates the identity on the manifold.\\[1pt]
\begin{figure*}[t]
\centering
\includegraphics[scale = 0.88]{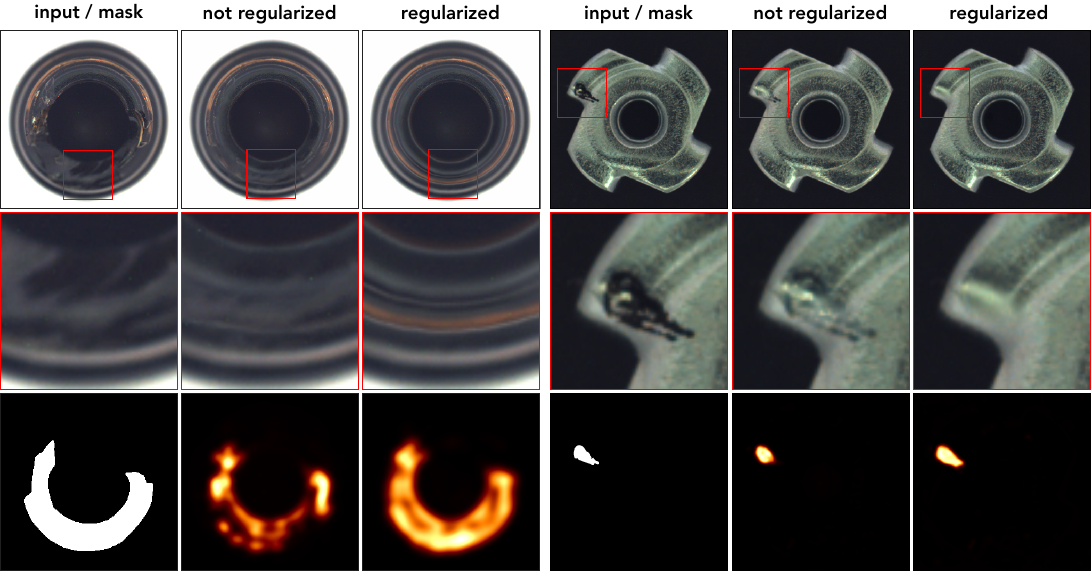}
\caption{
Comparison of reconstruction quality between our models trained with and without Gaussian noise. 
Each example shows a defective input $\hat{\bfx}$ (top row), its reconstructions $f(\hat{\bfx})$ from both models, and zoom-ins (middle row) of the region marked by a red square for detailed comparison. 
The bottom row shows the ground-truth anomaly mask and corresponding heatmaps. 
Reconstructions in corrupted regions and the corresponding anomaly heatmaps are visibly \textbf{improved} when the model is \textbf{regularized} with Gaussian noise.
}
\label{fig_recon_quality}
\end{figure*}
\textbf{Correction of anomalous regions.}
The Jacobian penalty enhances the correction of anomalous regions by balancing the influence of the reconstruction and Jacobian terms. 
While the reconstruction loss dominates in directions orthogonal to the manifold, enforcing projection back to it, 
the Jacobian term normalizes gradients along tangent directions, preventing \emph{mis-scaling}: 
if $J_{\bftheta}(\hat{\bfx}) \approx \alpha I_d$ with $\alpha < 1$, tangent variations are overly contracted and corruptions persist in smeared form, 
whereas for $\alpha > 1$ they are expanded and may be amplified. 
Only when $\alpha \approx 1$ do tangent directions behave like the identity, preserving normal structure while allowing orthogonal deviations to be corrected by the reconstruction term. 
This interaction compels the model to separate noise-like tangent perturbations from true structural anomalies, 
leading it to preserve normal variations while suppressing off-manifold deviations. 
Consequently, reconstructions become more stable and precise, as illustrated in Figure~\ref{fig_recon_quality}.

\subsection{Connection to DAE and RCAE}
An interesting question is how our regularization relates to earlier variants of regularized autoencoders that combine reconstruction and Jacobian penalties.
A prominent example is the Reconstruction Contractive Autoencoder (RCAE)~\cite{AlainB14}, defined as
\[
\mathcal{L}_{\text{RCAE}}(\bfx) = \|f_{\bftheta}(\bfx)-\bfx\|^2 + \lambda \|J_{\bftheta}(\bfx)\|_F^2 ,
\]
which minimizes a reconstruction loss on clean inputs together with a contractive penalty that drives the Jacobian toward zero. 
This induces an anisotropic effect: gradients remain non-zero along tangent directions for faithful reconstruction, 
but are suppressed in normal directions to reject off-manifold perturbations.
As shown in~\cite{AlainB14}, RCAE connects to the denoising autoencoder (DAE)~\cite{VincentLBM08}, 
since training with Gaussian noise implicitly adds a contractive Jacobian term (Bishop’s theorem).

Our approach differs in that it anchors the Jacobian to the identity, $J_{\bftheta}\!\approx\! I_d$, and evaluates it at corrupted inputs $\hat{\bfx}$.
Rather than enforcing flatness, this constraint promotes identity-like behavior along the manifold and stabilizes repairs off it.
In practice, this reduces the tendency toward high-frequency attenuation induced by strong Jacobian penalties, yielding sharper reconstructions and more reliable detection of structural anomalies.
Moreover, while RCAE requires explicit Jacobian computation, our formulation achieves the effect by simply adding Gaussian noise to inputs and outputs during training.

Finally, the asymptotic expansion in~(\ref{E_24092037}) shows that a fixed noise variance $\sigma^2$ sets the strength of the identity-anchoring term to $\sigma^2\|J_{\bftheta}(\hat{\bfx})-I_d\|_F^2$.
In practice, this biases the model toward the noise level seen during training. 
Varying $\sigma$ during training (e.g., $\sigma\!\sim\!\mathrm{Uniform}(0,r)$) mitigates this bias and improves robustness to noise-level variation.

\section{Corruption Model}
\label{sec:section4}
Training a standard autoencoder on normal samples alone yields poor anomaly detection performance, 
as the reconstruction objective is indifferent to whether inputs are normal or anomalous, 
rewarding only visual fidelity. 
Consequently, such models often reproduce normal and anomalous structures with comparable accuracy.
Reducing capacity through bottleneck compression or sparsity does not alter the objective and thus cannot resolve this limitation.

AD is inherently a binary classification task, 
yet its unsupervised formulation arises due to absence of labeled anomalous samples. 
We show that real anomalies are not strictly required: 
training on artificially corrupted data is sufficient when the corruption model 
captures essential aspects of structural deviations. 
Such distortions can be generated in a simple and fully self-supervised manner, 
enabling scalable and domain-agnostic training, as demonstrated in our experiments.

\begin{figure}[t]
\centering
\includegraphics[scale = 0.52]{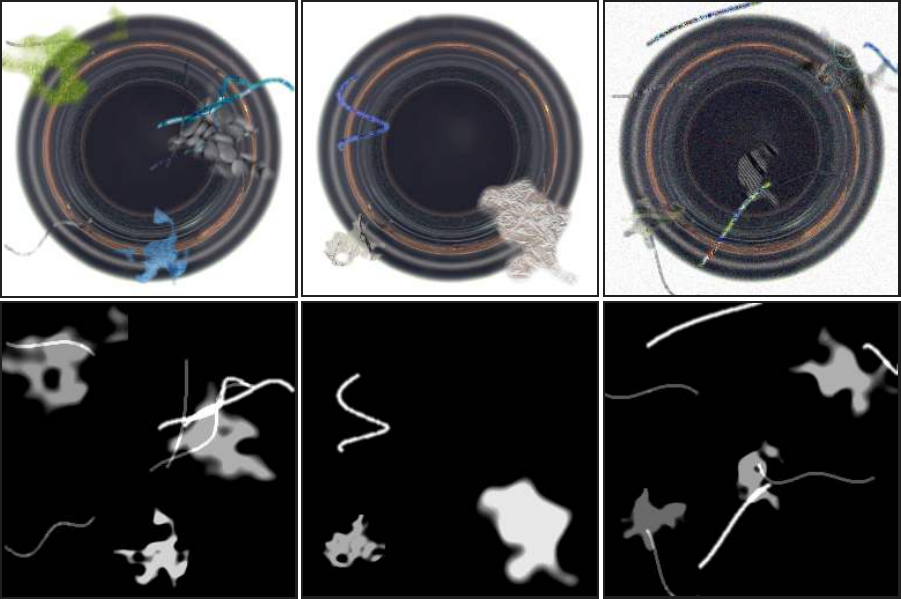}
\caption{Illustration of a combined realization of our design principles for simulating structural anomalies, 
covering variations in shape, texture, and occlusion opacity. 
The first row shows partially corrupted images, and the second row displays the corresponding 
anomaly masks, where mask intensity reflects the opacity of the occlusions. 
In the third example, Gaussian noise is applied across the entire image 
as part of our regularization strategy.}
\label{fig_bauer2}
\end{figure}

We distill prior approaches \cite{ZavrtanikKS212, pmlrkascenas22a, LiSYP21, 9412842, 10204486, math12243988} into a compact set of simple design principles 
enabling models trained on artificial corruptions to generalize to real defects.
We argue that artificial anomalies should be \emph{partial} 
and vary along three dimensions: \emph{shape}, \emph{texture}, and \emph{opacity}. 
Figure~\ref{fig_bauer2} illustrates their joint realization.

\subsection{Design Principles: Shape}
A crucial requirement is that the artificial occlusions used for training remain \emph{partial}, 
affecting only localized regions rather than the entire image. 
In a classical DAE, 
noise applied to the whole input often produces blurred or averaged reconstructions—even in uncorrupted regions. 
This occurs because the injected noise perturbs both orthogonal and tangent directions of the data manifold, 
causing the model to average over multiple plausible reconstructions. 
Partial distortions create a stronger contrast between normal and corrupted regions, 
allowing the model to focus on accurately reconstructing intact areas 
and thereby improving its discriminative ability between normal and anomalous regions.

\begin{remark}
Training with partial, localized occlusions of varying sizes—rather than global corruptions—preserves 
the fidelity of reconstructed normal regions while enhancing the model’s sensitivity to anomalous structures.
\end{remark}

In our implementation, occlusion shapes are generated by applying elastic deformation to ellipsoidal binary masks, 
following the approach of \cite{math12243988}. 
While this yields diverse shapes, they exhibit limited local geometric variation, 
with nearly constant cross-sectional thickness. 
To address this limitation, we superimpose additional curve-like structures of varying length, 
generated independently of the base mask. 
Combining these two extremes—broad, bulky forms and thin, elongated structures—proved sufficient to achieve strong performance. 
Notably, incorporating thin, curve-like masks substantially improved detection accuracy 
for several categories of the MVTec dataset. 
Figures~\ref{fig_bauer} in the supplement show qualitative improvements in the anomaly heatmaps for several examples.

\begin{remark}
\label{remark_2}
Occlusion masks should span the full spectrum of geometric variability, 
encompassing both diverse outer boundaries and internal shape characteristics—ranging from 
broad, bulky forms to thin, elongated shapes.
\end{remark}

\subsection{Design Principles: Texture Pattern}
In general, the range of texture variations used for corruption should be sufficiently diverse to ensure reliable detection of anomalous patterns. 
At the same time, artificial occlusions risk introducing non-natural artifacts that can lead to overfitting. 
A natural and effective way to mitigate this risk is to use real textures as anomaly fillers. 
In our implementation, we sample patches from the publicly available texture background dataset~\cite{cimpoi14describing} 
and insert them into training images according to generated shape masks. 
Optional post-processing steps, such as boundary smoothing or blending, further reduce artifacts 
while preserving control over the geometry of occlusions.

\begin{remark}
Artificial occlusions must balance diversity and naturalness:
rich texture variation promotes generalization, whereas adherence to natural image statistics prevents spurious artifacts.
\end{remark}

\subsection{Design Principles: Opacity}
Fully opaque occlusions maximally promote context-based reconstruction, 
as the model must infer the missing content solely from the surrounding regions. 
Reducing opacity, by contrast, partially reveals the underlying signal, 
encouraging more subtle correction mechanisms and acting as an implicit regularizer. 
Both effects are complementary: opaque occlusions emphasize global consistency, 
while transparent ones promote fine-grained fidelity. 
Therefore, a well-designed corruption model should include both. 

Moreover, real-world data support this distinction, as actual anomalies occur in both forms. 
For example, the \emph{bottle} category of the MVTec dataset contains contaminations that fully occlude the background, 
whereas the \emph{glue strip} class of the \emph{tile} category represents an example of partially transparent anomalies. 

\begin{remark}
A corruption model should incorporate both fully opaque and partially transparent occlusions. 
Opaque occlusions compel the model to reconstruct missing regions from context, 
whereas transparent ones regularize the learning process by enabling partial signal propagation.
\end{remark}

\section{Experiments}
\label{sec:section5}
\begin{figure*}[t]
\centering
\includegraphics[scale = 0.8]{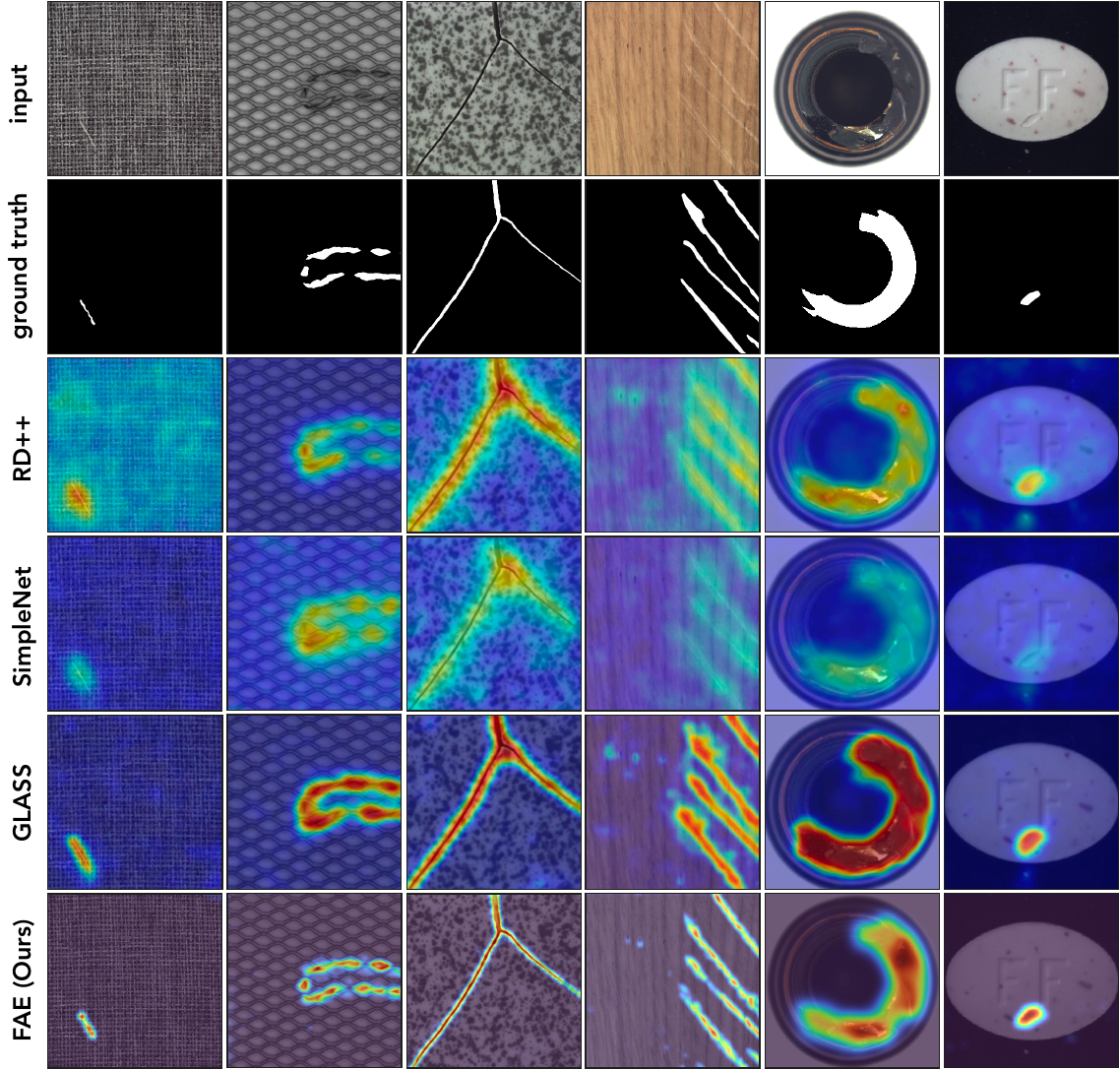}
\caption{
Qualitative comparison of anomaly heatmaps on representative samples from the MVTec AD dataset. Each column displays the input image, the corresponding human annotation mask, and heatmaps generated by multiple existing methods, with the results of our method shown at the bottom for comparison.
}
\label{fig_comparison}
\end{figure*}

\newcolumntype{g}{>{\columncolor{ColumnColor}}c}
\newcolumntype{q}{>{\columncolor{ColumnColor2}}c}
\begin{table*}[t]
  \caption{Experimental results for \textbf{anomaly recognition} measured with \textbf{I-AUROC} on the MVTec AD dataset.}
  \label{table_image_AUROC}
  \centering
  \scalebox{0.68}{
  \begin{tabular}{lqqqggggqqqqqqqqg}
    \toprule
    Category & AnoGAN & VAE  & LSR & RIAD & CutPaste & InTra & DRAEM & SimpleNet & PatchCore & RD++ & MSFlow & PNI & RealNet & SDC & GLASS & FAE \\
    \midrule
    & \multicolumn{3}{c}{\textit{2017 -- 2020}} & \multicolumn{4}{c}{\textit{2020 -- 2022}} & \multicolumn{8}{c}{\textit{2022 -- 2025}}  &  \multicolumn{1}{c}{\hspace*{5pt} \textit{Ours}\hspace*{5pt}}\\
    \midrule
    carpet  & 49 & 78 & 71 &  84.2 & 93.1 & 98.8 & 97.0 & 99.7 & 98.2 & \textcolor{red}{100} & \textcolor{red}{100} & \textcolor{red}{100} & \textcolor{blue}{99.8} & \textcolor{red}{100} & \textcolor{blue}{99.8} & \textcolor{red}{100}\\
    grid      & 51 & 73 & 91 & 99.6 & \textcolor{blue}{99.9} & \textcolor{red}{100} & \textcolor{blue}{99.9} & 99.7 & 98.3 & \textcolor{red}{100} & 99.8 & 98.4 & \textcolor{red}{100} & \textcolor{red}{100} & \textcolor{red}{100} & \textcolor{red}{100}\\
    leather & 52 & 95 & \textcolor{blue}{96} & \textcolor{red}{100} & \textcolor{red}{100} & \textcolor{red}{100} & \textcolor{red}{100} & \textcolor{red}{100} & \textcolor{red}{100} & \textcolor{red}{100} & \textcolor{red}{100} & \textcolor{red}{100} &\textcolor{red}{100} & \textcolor{red}{100} & \textcolor{red}{100} & \textcolor{red}{100}\\
    tile        & 51 & 80 & 95 &  93.4 & 93.4 & 98.2 & 99.6 & \textcolor{blue}{99.8} & 98.9 & 99.7 & \textcolor{red}{100} & \textcolor{red}{100} & \textcolor{red}{100} & \textcolor{red}{100} & \textcolor{red}{100} & \textcolor{red}{100}\\
    wood    & 68 & 77 & 96 & 93.0 & 98.6 & 97.5 & 99.1 & \textcolor{red}{100} & \textcolor{blue}{99.9} & 99.3 & \textcolor{red}{100} & 99.6 & 99.2 & \textcolor{red}{100} & \textcolor{blue}{99.9} & \textcolor{red}{100}\\
    \midrule
    avg. textures  & 54.2 & 80.6 & 89.8 & 94.0 & 97.0 & 98.9 & 99.1 & 99.8 & 99.0 & 99.8 & \textcolor{blue}{99.9} & 99.6 & 99.8 & \textcolor{red}{100} & \textcolor{blue}{99.9} & \textcolor{red}{100}\\
    \midrule
    bottle         & 69 & 87 & 99 & \textcolor{blue}{99.9} & 98.3 & \textcolor{red}{100} & 99.2 & \textcolor{red}{100} & \textcolor{red}{100} & \textcolor{red}{100} & \textcolor{red}{100} & \textcolor{red}{100} & \textcolor{red}{100} & \textcolor{red}{100} & \textcolor{red}{100} & \textcolor{red}{100}\\
    cable         & 53 & 90 & 72 & 81.9 & 80.6 & 70.3 & 91.8 & \textcolor{red}{99.9} & 99.7 & 99.2 & 99.5 & \textcolor{blue}{99.8} & 99.2 & \textcolor{red}{99.9} & \textcolor{blue}{99.8} & \textcolor{red}{99.9}\\
    capsule     & 58 & 74 & 68 & 88.4 & 96.2 & 86.5 & 98.5 & 97.7 & 98.1 & 99.0 & 99.2 & \textcolor{blue}{99.7} & 99.6 & 98.8 & \textcolor{red}{99.9} & \textcolor{red}{99.9}\\
    hazelnut    & 50 & 98 & 94 & 83.3 & \textcolor{blue}{97.3} & 95.7 & \textcolor{red}{100} & \textcolor{red}{100} & \textcolor{red}{100} & \textcolor{red}{100} & \textcolor{red}{100} & \textcolor{red}{100} & \textcolor{red}{100} & 98.8 & \textcolor{red}{100} & \textcolor{red}{100}\\
    metal nut   & 50 & 94 & 83 & 88.5 & 99.3 & 96.9 & 98.7 & \textcolor{red}{100} & \textcolor{red}{100} & \textcolor{red}{100} & \textcolor{red}{100} & \textcolor{red}{100} & \textcolor{blue}{99.8} & \textcolor{red}{100} & \textcolor{red}{100} & \textcolor{red}{100}\\
    pill             & 62 & 83 & 68 & 83.8 & 92.4 & 90.2 & 98.9 & 99.0 & 97.1 & 98.4 & \textcolor{blue}{99.6} & 96.9 & 99.1 & \textcolor{red}{100} & 99.4 & \textcolor{red}{100}\\
    screw        & 35 & 97 & 80 & 84.5 & 86.3 & 95.7 & 93.9 & 98.2 & 99.0 & 98.9 & 97.8 & 99.5 & 98.8 & \textcolor{blue}{99.6} & \textcolor{red}{100} & \textcolor{red}{100}\\
    toothbrush & 57 & 94 & 92 & \textcolor{red}{100} & 98.3 & \textcolor{red}{100} & \textcolor{red}{100} & \textcolor{blue}{99.7} & 98.9 & \textcolor{red}{100} & \textcolor{red}{100} & \textcolor{blue}{99.7} & 99.4 & 98.9 & \textcolor{red}{100} & \textcolor{red}{100}\\
    transistor   & 67 & 93 & 73 & 90.9 & 95.5 & 95.8 & 93.1 & \textcolor{red}{100} & 99.7 & 98.5 & \textcolor{red}{100} & \textcolor{red}{100} & \textcolor{red}{100} & \textcolor{red}{100} & \textcolor{blue}{99.9} & \textcolor{red}{100}\\
    zipper        & 59 & 78 & 97 & 98.1 & 99.4 & 99.4 & \textcolor{red}{100} & \textcolor{blue}{99.9} & 99.7 & 98.6 & \textcolor{red}{100} & \textcolor{blue}{99.9} & 99.8 & \textcolor{red}{100} & \textcolor{red}{100} & \textcolor{red}{100}\\
    \midrule
    avg. objects & 56.0 & 88.8 & 82.6 & 89.9 & 94.4 & 93.1 & 97.4 & 99.5 & 99.2 & 99.3 & 99.6 & 99.5 & 99.6 &  \textcolor{blue}{99.7} & \textcolor{red}{99.9} & \textcolor{red}{99.9}\\
    \midrule
    avg. all        &  55.4 & 86.1 & 85.0 & 91.3 & 95.2 & 95.0 &  98.0  & 99.6 & 99.2 & 99.4 & 99.7 & 99.6 & 99.7 &  \textcolor{blue}{99.8} & \textcolor{red}{99.9} & \textcolor{red}{99.9}\\
    \bottomrule
  \end{tabular}}
\end{table*}
\begin{table*}[t]
  \caption{Experimental results for \textbf{anomaly segmentation} measured with \textbf{P-AUROC} on the MVTec AD dataset.}
  \label{table_AUROC}
  \centering
  \scalebox{0.68}{
  \begin{tabular}{lqqqggggqqqqqqqqg}
    \toprule
    Category & AnoGAN & VAE & LSR  & RIAD & CutPaste & InTra & DRAEM & SimpleNet & PatchCore & RD++ & MSFlow & PNI & RealNet & SDC & GLASS & FAE\\
    \midrule
    & \multicolumn{3}{c}{\textit{2017 -- 2020}} & \multicolumn{4}{c}{\textit{2020 -- 2022}} & \multicolumn{8}{c}{\textit{2022 -- 2025}}  &  \multicolumn{1}{c}{\hspace*{5pt} \textit{Ours}\hspace*{5pt}}\\
    \midrule
    carpet  &  54  &  78  & 94 & 96.3 & 98.3 & 99.2 & 95.5 & 98.2 & 98.7 & 99.2 & 99.4 & 99.4 & 99.2 & \textcolor{red}{99.8} & \textcolor{blue}{99.6} & \textcolor{red}{99.8}\\
    grid &       58  &  73  & 99 & 98.8  & 97.5 & 98.8 & \textcolor{blue}{99.7} & 98.8 & 98.8 & 99.3 & 99.4 & 99.2 & 99.5 & \textcolor{red}{99.8} & 99.4 & \textcolor{red}{99.8}\\
    leather &  64 &  95 & 99 & 99.4 & 99.5 & 99.5 & 98.6 & 99.2 & 99.3 & 99.4 & \textcolor{blue}{99.7} & 99.6 & \textcolor{red}{99.8} & \textcolor{blue}{99.7} & \textcolor{red}{99.8} & \textcolor{red}{99.8}\\
    tile &         50 &  80  & 88 & 89.1 &  90.5 & 94.4 & 99.2 & 97.0 & 96.3 & 96.6 & 98.2 & 98.4 & \textcolor{blue}{99.4} & 99.2 & \textcolor{red}{99.7} & \textcolor{blue}{99.4}\\
    wood &     62  & 77  & 87 & 85.8 & 95.5 & 88.7 & 96.4 & 94.5 & 95.2 & 95.8 & 97.1 & 97.0 & 98.2 & 98.4 & \textcolor{red}{98.8} & \textcolor{blue}{98.6}\\
    \midrule
    avg. tex.  & 57.6 & 80.6 & 93.4 & 93.9 & 96.3 & 96.1 & 97.9 & 97.5 & 97.7 & 98.1 & 98.8 & 98.7 & 99.2 & \textcolor{blue}{99.4} & \textcolor{red}{99.5} & \textcolor{red}{99.5}\\
    \midrule
    bottle &  86  & 87 & 95 & 98.4 & 97.6 & 97.1 & \textcolor{blue}{99.1} & 98.0 & 98.6 & 98.8 & 99.0 & 98.9 & \textcolor{red}{99.3} & 98.9 & \textcolor{red}{99.3} & \textcolor{red}{99.3}\\
    cable &  86  & 87 & 95 & 94.2 & 90.0 & 91.0 & 94.7 & 97.6 & \textcolor{blue}{98.7} & 98.4 & 98.5 & \textcolor{red}{99.1} & 98.1 & 98.5 & \textcolor{blue}{98.7} & \textcolor{blue}{98.7}\\
    capsule &  84  & 74 & 93 & 92.8 & 97.4 & 97.7 & 94.3 & 98.9 & 99.1 & 98.8 & 99.1 & \textcolor{blue}{99.3} & \textcolor{blue}{99.3} & 99.1 & \textcolor{red}{99.4} & \textcolor{blue}{99.3}\\
    hazelnut & 87  & 98 & 95 & 96.1 & 97.3 & 98.3 & \textcolor{red}{99.7} & 97.9 & 98.8 & 99.2 & 98.7 & 99.4 & \textcolor{red}{99.7} & 99.1 & 99.4 & \textcolor{blue}{99.6}\\
    metal nut &  76  & 94 & 91 & 92.5 & 93.1 & 93.3 & \textcolor{red}{99.5} & 98.8 & 99.0 & 98.1 & 99.3 & 99.3 & 98.6 & 98.5 & \textcolor{blue}{99.4} & \textcolor{red}{99.5}\\
    pill & 87  & 83 & 91 & 95.7 & 95.7 & 98.3 & 97.6 & 98.6 & 98.6 & 98.3 & 98.8 & 99.0 & 99.0 & \textcolor{blue}{99.3} & \textcolor{red}{99.5} & \textcolor{blue}{99.3}\\
    screw  & 80 & 97 & 96 & 98.8 & 96.7 & 99.5 & 97.6 & 99.3 & 99.5 & \textcolor{red}{99.7} & 99.1 & \textcolor{blue}{99.6} & 99.5 & \textcolor{red}{99.7} & 99.5 & \textcolor{red}{99.7}\\
    toothbrush & 90 & 94 & 97 & 98.9 & 98.1 & 98.9 & 98.1 & 98.5 & 98.9 & 99.1 & 98.5 & 99.1 & 98.7 & \textcolor{blue}{99.4} & 99.3 & \textcolor{red}{99.5}\\
    transistor &  80 & 93 & 91 & 87.7 & 93.0 & 96.1 & 90.9 & 97.6 & 97.1 & 0.0 & \textcolor{blue}{98.3} & 98.0 & 98.0 & \textcolor{red}{98.9} & 97.6 & \textcolor{red}{98.9}\\
    zipper &  78 & 78 & 98 & 97.8 & 99.3 & 99.2 & 98.8 & 98.9 & 99.0 & 94.3 & 99.2 & 99.4 & 99.2 & \textcolor{blue}{99.6} & \textcolor{blue}{99.6} & \textcolor{red}{99.7}\\
    \midrule
    avg. obj. & 83.4 & 88.5 & 94.6 & 95.3 & 95.8 & 96.9 & 97.0 & 98.4 & 98.7 & 98.8 & 98.8 & 99.1 & 98.9 & 99.1 & \textcolor{blue}{99.2} & \textcolor{red}{99.4} \\
    \midrule
    avg. all & 74.8 & 85.9 & 94.2 & 94.8 & 96.0 & 96.7 & 97.3 & 98.1 & 98.4 & 98.3 & 98.8 & 99.0 & 99.0 & 99.2 & \textcolor{blue}{99.3} & \textbf{\textcolor{red}{99.4}} \\
    \bottomrule
  \end{tabular}}
\end{table*}
\begin{figure*}[t]
\centering
\includegraphics[scale = 0.81]{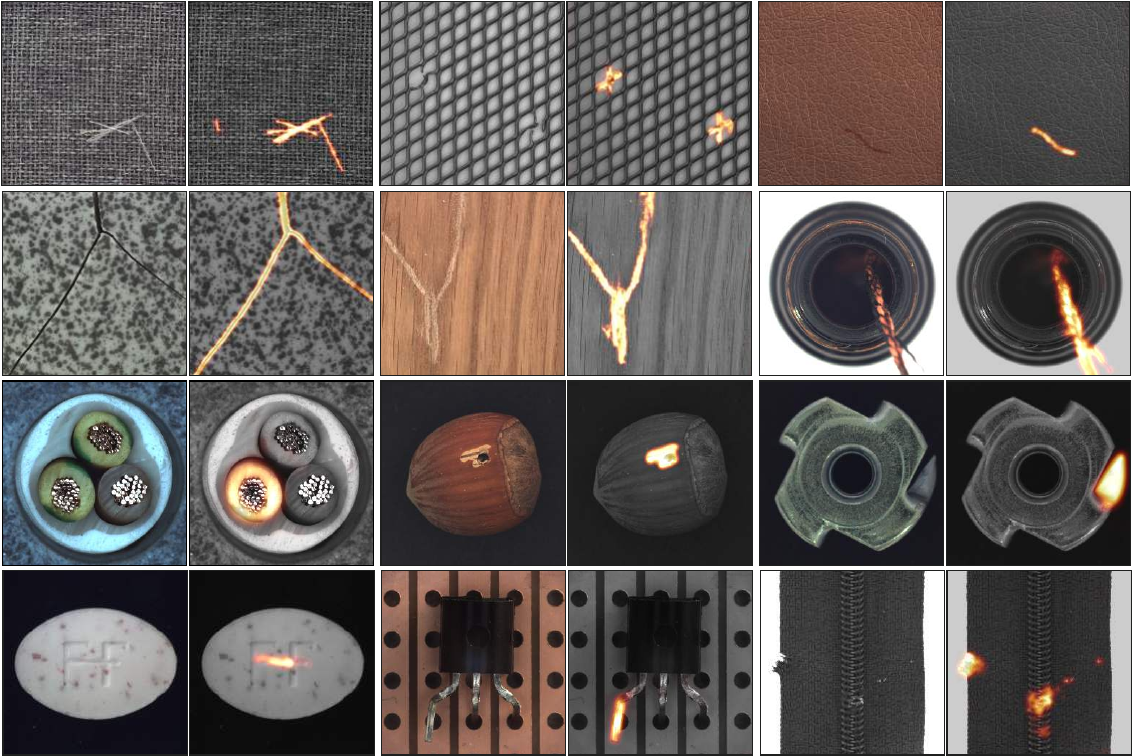}
\caption{
Representative (randomly chosen) examples of anomaly heatmaps on the MVTec~AD dataset produced by our model. Each pair shows an input image and its corresponding heatmap overlay highlighting the detected defective regions.
}
\label{fig_bauer4}
\end{figure*}
In this section, we present results demonstrating the effectiveness of our approach for robust structural anomaly detection. 
We train a Filtering Autoencoder (FAE) using our corruption model, regularized to preserve Gaussian noise.\\
\textbf{Dataset.} We evaluate on the widely adopted MVTec AD benchmark \cite{BergmannFSS19, BergmannBFSS21}, 
which contains 5,354 high-resolution images across 15 categories (10 objects and 5 textures) with substantial variability in defect type and appearance. 
The training set comprises only defect-free images, while the test set includes both defect-free and anomalous images. 
There are over 70 different types of anomalies including scratches, dents, contaminations, and structural changes (in total ~1,888 manually annotated defect regions).\\ 
\textbf{Training setup.} To increase the number of training samples, we apply category-dependent augmentations 
such as random rotations and flips, holding out 5\% of the training data for validation. 
For the \emph{transistor} class, additional $90^\circ$ rotations
were essential to detect missing (or rotated) components, although this made training more challenging.  
We train one model per category using a modified U-Net with dilated convolutions in the bottleneck. 
Training is performed with the Adam optimizer \cite{KingmaB14} at a learning rate of $10^{-4}$. 
Texture categories are resized to $512 \times 512$, while object categories are downscaled to $256 \times 256$ pixels.\\
\textbf{Results.}
We report image-level (I-AUROC) and pixel-level (P-AUROC) performance in 
Tables~\ref{table_image_AUROC} and~\ref{table_AUROC}, respectively. 
Following established convention, the values are rounded to one decimal place 
for compatibility with prior work. 
For the detection task, our method achieves a total average score of  99.9. 
For the segmentation task, we obtain a new state-of-the-art result of 99.4, 
surpassing a range of top-performing methods. 
We compare against a selection of established and leading baselines, including 
AnoGAN~\cite{SchleglSWSL17}, 
VAE~\cite{LiuLZKWBRC20}, 
LSR~\cite{LSR}, 
RIAD~\cite{ZavrtanikKS21}, 
CutPaste~\cite{LiSYP21}, 
InTra~\cite{InTra}, 
DRAEM~\cite{ZavrtanikKS212}, 
SimpleNet~\cite{LiuZXW23}, 
PatchCore~\cite{RothPZSBG22}, 
RD++~\cite{10204486}
MSFlow~\cite{msflow-abs-2308-15300}, 
PNI~\cite{Jaehyeok2023}, 
RealNet~\cite{10658311}, 
SDC~\cite{math12243988}, 
and GLASS~\cite{ChenLLZ24}. 
All comparison scores are taken directly from the respective publications.
To illustrate the quality and sharpness of the resulting anomaly heatmaps produced by our model, representative examples are shown in Figure~\ref{fig_bauer4}. 
Supplementary material provides additional examples of reconstruction improvements (Figure~\ref{fig_bauer5}) and resulting anomaly heatmaps (Figures~\ref{fig_examples} and~\ref{fig_examples2}),
as well as a direct comparison of heatmap quality with previous methods (Figure~\ref{fig_comparison}).

\section{Conclusion}
\label{sec:section6}
We proposed the novel \emph{Filtering Autoencoder} (FAE) for robust structural anomaly detection,  
trained to correct artificially corrupted samples.  
Unlike conventional denoising autoencoders, the FAE preserves Gaussian noise,  
which acts as a Tikhonov regularizer anchoring the Jacobian toward the identity.  
This stabilizes both reconstruction and anomaly detection,  
allowing the model to disentangle structured deviations from random fluctuations.  
Meanwhile, the regularization remains lightweight and method-agnostic,  
adding negligible computational overhead and integrating seamlessly with other restoration-based approaches.

We further proposed a generic corruption-model template that (together with noise regularization)
effectively mitigates overfitting to artificial corruption patterns and enhances generalization to real anomalies.
On the MVTec~AD benchmark, our approach achieves 99.4~P-AUROC, surpassing prior methods.
The robustness and accuracy of the FAE make it a strong candidate for industrial inspection.

While effective for structural anomalies, reconstruction-based models alone cannot address  
logical anomalies involving ambiguous or ill-defined annotations.  
Future work will explore integrating our approach with constraint optimization~\cite{math11122628, BauerSM17, Bauer2019}  
and reasoning-driven methods~\cite{ZhuCDOW24, XuL0PD25}.

\raggedbottom

\section*{Acknowledgements}
This work was supported by the German Federal Ministry of Research, Technology and Space (BMFTR) under grant Nr. BIFOLD25B. K.-R.M. was partly supported by the Institute of Information \& Communications Technology Planning \& Evaluation (IITP) grant funded by the Korea government (MSIT) (No. RS-2019-II190079, Artificial Intelligence Graduate School Program, Korea University) and grant funded by the Korea government (MSIT) (No. RS-2024-00457882, AI Research Hub Project).
{
    \small
    \bibliographystyle{ieeenat_fullname}
    \bibliography{references}
}

 \clearpage
\setcounter{page}{1}
\maketitlesupplementary

\section{Proof of Theorem \ref{thm:ia_tikhonov_pointwise_general}}
We first prove the following proposition.
\begin{proposition}
\label{prop_17101821}
Let $d \in \mathbb{N}$, $\hat{\bfx},\bfx \in \mathbb{R}^d$ and 
$\boldsymbol{\epsilon}\sim\mathcal N(\mathbf 0,\sigma^2 I_d)$.  
Assume $f_{\bftheta} \in C^4$ in a neighborhood of $\hat{\bfx}$ and
define
\begin{equation}
\mathcal{L}(\hat{\bfx},\bfx,\sigma; \bftheta)
:= \mathbb{E}_{\boldsymbol{\epsilon}}
\Bigl[\|\,f_{\bftheta}(\hat{\bfx}+\boldsymbol{\epsilon})-(\bfx+\boldsymbol{\epsilon})\,\|^2_2\Bigr].
\end{equation}
Then, as $\sigma \to 0$, the loss $\mathcal{L}(\hat{\bfx},\bfx,\sigma; \bftheta)$ admits the asymptotic expansion
\begin{equation}
\| \bfr \|^2_2 + \sigma^2 \big( \| J_{\bftheta}(\hat{\bfx})-I_d \|_F^2 + \bfr^\top \Delta f_{\bftheta}(\hat{\bfx}) \big)
+ \mathcal{O}(\sigma^4),
\end{equation}
where $\bfr = f_{\bftheta}(\hat{\bfx})-\bfx$ is the residual, $J_{\bftheta}(\hat{\bfx})=D_{\hat{\bfx}}f_{\bftheta}$ is the Jacobian, 
$\Delta f_{\bftheta}$ is the component-wise Laplacian of $f_{\bftheta}$,
and $I_d$ denotes the identity matrix.
\end{proposition}

\noindent Let $\mathbf r := f_{\boldsymbol\theta}(\hat{\bfx})-\bfx$, 
$J := D_{\hat{\bfx}} f_{\boldsymbol\theta}$, and
$\boldsymbol\epsilon\sim\mathcal N(\mathbf 0,\sigma^2 I_d)$.
\paragraph{Step 1: Second-order Taylor expansion.}
Since $f_{\boldsymbol\theta}$ is $C^3$ in a neighborhood of $\hat{\bfx}$, Taylor’s theorem (Fréchet form) gives
\[
f_{\boldsymbol\theta}(\hat{\bfx}+\boldsymbol\epsilon)
= f_{\boldsymbol\theta}(\hat{\bfx}) + J\,\boldsymbol\epsilon 
+ \tfrac12\,D^2_{\hat{\bfx}} f_{\boldsymbol\theta}(\boldsymbol\epsilon,\boldsymbol\epsilon)
+ R(\hat{\bfx}, \boldsymbol\epsilon),
\]
where, for some $t \in (0,1)$,
\begin{equation}
\label{E_28091121}
R(\hat{\bfx}, \boldsymbol\epsilon)
= \tfrac16\,D^3_{\hat{\bfx} + t \boldsymbol\epsilon} f_{\boldsymbol\theta}
  (\boldsymbol\epsilon,\boldsymbol\epsilon,\boldsymbol\epsilon).
\end{equation}
Since $D^3 f_{\boldsymbol\theta}$ is continuous in a neighborhood of $\hat{\bfx}$, it is locally bounded; hence there exist $\rho>0$ and $C>0$ such that
$\|D^3_{\bfy} f_{\boldsymbol\theta}\|\le C$ for all $\bfy$ with $\|\bfy-\hat{\bfx}\|_2\le \rho$.
For any $\boldsymbol{\epsilon}$ with $\|\boldsymbol{\epsilon}\|_2\le \rho$ we get
\begin{equation}
\big\|R(\hat{\bfx},\boldsymbol{\epsilon})\big\|_2
\overset{(\ref{E_28091121})}{=} \tfrac16 \big\|D^3_{\hat{\bfx}+t\boldsymbol{\epsilon}} f_{\boldsymbol\theta} (\boldsymbol{\epsilon},\boldsymbol{\epsilon},\boldsymbol{\epsilon})\big\|_2 \le \tfrac{C}{6}\,\|\boldsymbol{\epsilon}\|_2^{3},
\end{equation}
where in the last step we used the operator norm of the trilinear map
\[
\|D^3_{y} f_{\boldsymbol\theta}(\boldsymbol\epsilon,\boldsymbol\epsilon,\boldsymbol\epsilon)\|_2 \le \|D^3_{y} f_{\boldsymbol\theta}\|\,\|\boldsymbol\epsilon\|_2^3 \le C \|\boldsymbol\epsilon\|_2^3.
\]
Set $M:=J-I_d$ and $A:=D^2_{\hat{\bfx}} f_{\boldsymbol\theta}$. Then
\[
g(\boldsymbol\epsilon)
:= f_{\boldsymbol\theta}(\hat{\bfx}+\boldsymbol\epsilon)-(\bfx+\boldsymbol\epsilon)
= \mathbf r + M\boldsymbol\epsilon + \tfrac12\,A(\boldsymbol\epsilon,\boldsymbol\epsilon)
+ R(\hat{\bfx}, \boldsymbol\epsilon).
\]

\paragraph{Step 2: Expand $\|g(\boldsymbol\epsilon)\|^2$ and take expectations.}
Expanding,
\begin{align*}
\|g(\boldsymbol\epsilon)\|_2^2
&= \|\mathbf r\|_2^2
+ \|M\boldsymbol\epsilon\|_2^2
+ \tfrac14\,\|A(\boldsymbol\epsilon,\boldsymbol\epsilon)\|_2^2
+ \|R(\hat{\bfx}, \boldsymbol\epsilon)\|_2^2 \\
&\quad
+ 2\,\mathbf r^\top (M\boldsymbol\epsilon)
+ \mathbf r^\top A(\boldsymbol\epsilon,\boldsymbol\epsilon)
+ 2\,\mathbf r^\top R(\hat{\bfx}, \boldsymbol\epsilon) \\
&\quad
+ \boldsymbol\epsilon^\top M^\top A(\boldsymbol\epsilon,\boldsymbol\epsilon)
+ 2\,\boldsymbol\epsilon^\top M^\top R(\hat{\bfx}, \boldsymbol\epsilon) \\
&\quad
+ A(\boldsymbol\epsilon,\boldsymbol\epsilon)^\top R(\hat{\bfx}, \boldsymbol\epsilon).
\end{align*}
Taking expectations and using $\mathbb E[\boldsymbol\epsilon]=0$, $\mathbb E[\boldsymbol\epsilon\boldsymbol\epsilon^\top]=\sigma^2 I_d$,
we obtain
\begin{equation*}
\begin{aligned}
\mathbb E[\mathbf r^\top (M\boldsymbol\epsilon)] &= 0, \\
\mathbb E\|M\boldsymbol\epsilon\|_2^2 &= \mathbb E[(M^\top M \boldsymbol\epsilon)^\top \boldsymbol\epsilon] = \mathbb E[\text{tr}((M^\top M \boldsymbol\epsilon)^\top \boldsymbol\epsilon)] \\
&= \mathbb E[\text{tr}(M^\top M \boldsymbol\epsilon \boldsymbol\epsilon^\top)] = \text{tr}(M^\top M \mathbb{E}[\boldsymbol\epsilon \boldsymbol\epsilon^\top]) \\
&= \sigma^2 \text{tr}(M^\top M) = \sigma^2\|M\|_F^2 \\
\mathbb E\!\left[\mathbf r^\top A(\boldsymbol\epsilon,\boldsymbol\epsilon)\right] &= \mathbf r^\top \mathbb E\!\left[A(\boldsymbol\epsilon,\boldsymbol\epsilon)\right] = \mathbf r^\top \mathbb{E}\!\left[ ((H_k \boldsymbol\epsilon)^\top \boldsymbol\epsilon)_{k=1}^d\right] \\
&= \mathbf r^\top \mathbb{E}\!\left[ (\text{tr}((H_k \boldsymbol\epsilon)^\top \boldsymbol\epsilon))_{k=1}^d\right] \\ 
&= \mathbf r^\top \mathbb{E}\!\left[ (\text{tr}(H_k \boldsymbol\epsilon \boldsymbol\epsilon^\top))_{k=1}^d\right] \\
&= \mathbf r^\top (\text{tr}(H_k \mathbb{E}\!\left[\boldsymbol\epsilon \boldsymbol\epsilon^\top \right] ))_{k=1}^d \\
&= \sigma^2 \mathbf r^\top (\text{tr}(H_k))_{k=1}^d = \sigma^2\,\mathbf r^\top \Delta f_{\boldsymbol\theta}(\hat{\bfx}),
\end{aligned}
\end{equation*}
where $H_k$ is the Hessian of the $k$-th component of $f_{\bftheta}$.

\noindent Let $\phi(\boldsymbol\epsilon):=\boldsymbol\epsilon^\top M^\top A(\boldsymbol\epsilon,\boldsymbol\epsilon)$.
Then
\[
\mathbb E\big[\boldsymbol\epsilon^\top M^\top A(\boldsymbol\epsilon,\boldsymbol\epsilon)\big]
= \mathbb E[\phi(\boldsymbol\epsilon)]
= \mathbb E[\phi(-\boldsymbol\epsilon)]
= -\,\mathbb E[\phi(\boldsymbol\epsilon)]
\]
This implies:
\begin{equation}
\mathbb E\big[\boldsymbol\epsilon^\top M^\top A(\boldsymbol\epsilon,\boldsymbol\epsilon)\big] = 0.
\end{equation}
Now recall
\begin{equation}
\begin{aligned}
R(\hat{\bfx},\boldsymbol{\epsilon}) &= \tfrac16\,D^3_{\hat{\bfx}+ \theta \boldsymbol{\epsilon}} f_{\boldsymbol\theta} (\boldsymbol{\epsilon},\boldsymbol{\epsilon},\boldsymbol{\epsilon}), \\
R(\hat{\bfx},-\boldsymbol{\epsilon}) &= -\,\tfrac16\,D^3_{\hat{\bfx}- \theta' \boldsymbol{\epsilon}} f_{\boldsymbol\theta}
(\boldsymbol{\epsilon},\boldsymbol{\epsilon},\boldsymbol{\epsilon}),
\end{aligned}
\end{equation}
for some $\theta,\theta'\in(0,1)$. Hence
\[
R(\hat{\bfx},\boldsymbol{\epsilon})+R(\hat{\bfx},-\boldsymbol{\epsilon})
= \tfrac16\Big(D^3_{\hat{\bfx}+\theta \boldsymbol{\epsilon}} f_{\boldsymbol\theta}
               -D^3_{\hat{\bfx}-\theta' \boldsymbol{\epsilon}} f_{\boldsymbol\theta}\Big)
  (\boldsymbol{\epsilon},\boldsymbol{\epsilon},\boldsymbol{\epsilon}).
\]
Since $f_{\boldsymbol\theta}\in C^4$ near $\hat{\bfx}$, $D^4 f_{\boldsymbol\theta}$ is continuous and therefore bounded on a small closed ball $B$ around $\hat{\bfx}$; set
\[
L := \sup_{y\in B}\big\|D^4_{y} f_{\boldsymbol\theta}\big\| < \infty.
\]
By the mean-value inequality for the map $\bfy \mapsto D^3_{\bfy} f_{\boldsymbol\theta}$,
\begin{equation*}
\begin{aligned}
\big\|D^3_{\hat{\bfx}+\theta \boldsymbol{\epsilon}} f_{\boldsymbol\theta} - D^3_{\hat{\bfx}-\theta' \boldsymbol{\epsilon}} f_{\boldsymbol\theta}\big\|
&\le L\,\|\hat{\bfx}+\theta \boldsymbol{\epsilon}-(\hat{\bfx}-\theta' \boldsymbol{\epsilon})\|_2 \\
&= L\,(\theta+\theta')\,\|\boldsymbol{\epsilon}\|_2 .
\end{aligned}
\end{equation*}
Using the operator norm of a trilinear map,
\[
\big\|T(\boldsymbol{\epsilon},\boldsymbol{\epsilon},\boldsymbol{\epsilon})\big\|_2
\le \|T\|\,\|\boldsymbol{\epsilon}\|_2^3,
\]
for $T = D^3_{\hat{\bfx}+\theta \boldsymbol{\epsilon}} f_{\boldsymbol\theta}
     -D^3_{\hat{\bfx}-\theta' \boldsymbol{\epsilon}} f_{\boldsymbol\theta}$ we obtain
\begin{equation}
\label{E_28091621}
\begin{aligned}
\big\|R(\hat{\bfx},\boldsymbol{\epsilon})+R(\hat{\bfx},-\boldsymbol{\epsilon})\big\|_2
&\le \tfrac16\,L\,(\theta+\theta')\,\|\boldsymbol{\epsilon}\|_2^4 \\
&\le \tfrac{L}{3}\,\|\boldsymbol{\epsilon}\|_2^4.
\end{aligned}
\end{equation}
By symmetrization,
\[
\mathbb E\big[\mathbf r^\top R(\hat{\bfx},\boldsymbol{\epsilon})\big]
= \frac{1}{2}\,\mathbb E\!\left[\mathbf r^\top\big(R(\hat{\bfx},\boldsymbol{\epsilon})
+R(\hat{\bfx},-\boldsymbol{\epsilon})\big)\right].
\]
Using the pointwise Cauchy–Schwarz inequality $\,\mathbf r^\top v \le \|\mathbf r\|_2\,\|v\|_2$,
\[
\mathbb E\big[\mathbf r^\top R(\hat{\bfx},\boldsymbol{\epsilon})\big]
\le \frac{\|\mathbf r\|_2}{2}\,\mathbb E\big\|R(\hat{\bfx},\boldsymbol{\epsilon})
+R(\hat{\bfx},-\boldsymbol{\epsilon})\big\|_2.
\]
Combining with (\ref{E_28091621}) and Jensen inequality we get
\[
\mathbb E\big[\mathbf r^\top R(\hat{\bfx},\boldsymbol{\epsilon})\big]
\le \frac{\|\mathbf r\|_2}{2}\cdot \frac{L}{3}\,\mathbb E\|\boldsymbol{\epsilon}\|_2^4
= \frac{\|\mathbf r\|_2\,L}{6}\,(d^2+2d)\,\sigma^4
\]
since $\mathbb E\|\boldsymbol{\epsilon}\|_2^4=(d^2+2d)\sigma^4$ for $\boldsymbol{\epsilon}\sim\mathcal N(0,\sigma^2 I_d)$.
Therefore, 
\begin{equation}
\mathbb E\big[\mathbf r^\top R(\hat{\bfx},\boldsymbol{\epsilon})\big] \in \mathcal O(\sigma^4).
\end{equation}

\noindent \emph{Remaining terms.}
Let $A(\boldsymbol\epsilon,\boldsymbol\epsilon)
=(\boldsymbol\epsilon^\top H_1\boldsymbol\epsilon,\dots,\boldsymbol\epsilon^\top H_d\boldsymbol\epsilon)$ with $H_k$ being the Hessian of the $k$-th component of $f_{\bftheta}$.
Since $|\boldsymbol\epsilon^\top H_k\boldsymbol\epsilon|\le \|H_k\|_F\,\|\boldsymbol\epsilon\|_2^2$,
\begin{equation*}
\tfrac14\,\mathbb E\|A(\boldsymbol\epsilon,\boldsymbol\epsilon)\|_2^2
\in \mathcal O(\sigma^4), \hspace*{3pt}
\mathbb E\|R(\hat{\bfx}, \boldsymbol\epsilon)\|_2^2 \in \mathcal O(\sigma^6),
\end{equation*}
and by Cauchy–Schwarz,
\begin{equation*}
\mathbb E[\boldsymbol\epsilon^\top M^\top R(\hat{\bfx}, \boldsymbol\epsilon)] \in \mathcal O(\sigma^4), \hspace*{3pt}
\mathbb E\big[A(\boldsymbol\epsilon,\boldsymbol\epsilon)^\top R(\hat{\bfx}, \boldsymbol\epsilon)\big] \in \mathcal O(\sigma^5).
\end{equation*}

\paragraph{Step 3: Collecting terms.}
Keeping contributions up to order $\sigma^4$,
we can write $\mathcal{L}(\hat{\bfx},\bfx,\sigma; \bftheta)$ as
\[
\|\mathbf r\|_2^2
+ \sigma^2 \|J-I_d\|_F^2
+ \sigma^2\,\mathbf r^\top \Delta f_{\boldsymbol\theta}(\hat{\bfx})
+ \mathcal O(\sigma^4),
\]
which is the claimed expansion.\\[3pt]

\paragraph{Proof of Theorem \ref{thm:ia_tikhonov_pointwise_general}}.
We now prove Theorem \ref{thm:ia_tikhonov_pointwise_general} by using the result in Proposition \ref{prop_17101821}.
By independence of $(\hat{\bfx}, \bfx)$ and $\boldsymbol{\epsilon}$ we can write
\[
\mathcal{L}_{\mathrm{FAE}}(\sigma;\bftheta)
= \mathbb{E}_{(\hat{\bfx},\bfx)}\!\left[ 
\mathcal{L}(\hat{\bfx},\bfx,\sigma;\bftheta)\right],
\]
where
\[
\mathcal{L}(\hat{\bfx},\bfx,\sigma;\bftheta)
:= \mathbb{E}_{\boldsymbol{\epsilon}}
\left[\|f_{\bftheta}(\hat{\bfx}+\boldsymbol{\epsilon})
-(\bfx+\boldsymbol{\epsilon})\|^2_2\right].
\]

For fixed $(\hat{\bfx},\bfx)$, Theorem~\ref{thm:ia_tikhonov_pointwise_general} gives the following expansion
for $\mathcal{L}(\hat{\bfx},\bfx,\sigma;\bftheta)$
\[
\|\mathbf r\|_2^2
+ \sigma^2\!\left(
   \|J_{\bftheta}(\hat{\bfx})-I_d\|_F^2
   + r^\top \Delta f_{\bftheta}(\hat{\bfx})
 \right)
+ R_\sigma(\hat{\bfx}),
\]
where $\mathbf r = f_{\bftheta}(\hat{\bfx}) - \bfx$ and the remainder satisfies
\[
|R_\sigma(\hat{\bfx})| \le C(\hat{\bfx})\,\sigma^4,
\]
for some measurable $C(\hat{\bfx})$.
Since $f_{\bftheta}\in C^3$ and $D^3 f_{\bftheta}$ is locally Lipschitz, such a bound follows from the Taylor expansion with Gaussian averaging.
Taking expectations with respect to $(\hat{\bfx},\bfx)$ yields
\[
\mathbb{E}\!\left[
   \|\mathbf r\|_2^2
   + \sigma^2\!\left(
       \|J_{\bftheta}(\hat{\bfx})-I_d\|_F^2
       + r^\top \Delta f_{\bftheta}(\hat{\bfx})
     \right)\right]
+ \mathbb{E}[R_\sigma(\hat{\bfx})].
\]
By the integrability assumption
\[
\mathbb{E}\|\mathbf r\|_2^2, \mathbb{E}\|J_{\bftheta}(\hat{\bfx})\|_F^2, \mathbb{E}\|D^2_{\hat{\bfx}} f_{\bftheta}\|_F^2 < \infty,
\]
we also have $\mathbb{E}[C(\hat{\bfx})]<\infty$,
so $\mathbb{E}[R_\sigma(\hat{\bfx})]=\mathcal{O}(\sigma^4)$.
Thus, the loss $\mathcal{L}_{\mathrm{FAE}}(\sigma;\bftheta)$ is equal to
\[
\mathbb{E}\!\left[
   \|\mathbf r\|^2_2
   + \sigma^2\!\left(
       \|J_{\bftheta}(\hat{\bfx})-I_d\|_F^2
       + r^\top \Delta f_{\bftheta}(\hat{\bfx})
     \right)\right]
+ \mathcal{O}(\sigma^4).
\]
which is the claimed expansion. \hfill $\square$

\section{Varying Local Thickness of Occluding Shapes Improves Segmentation Accuracy}
\begin{figure}[t]
\centering
\includegraphics[scale = 0.45]{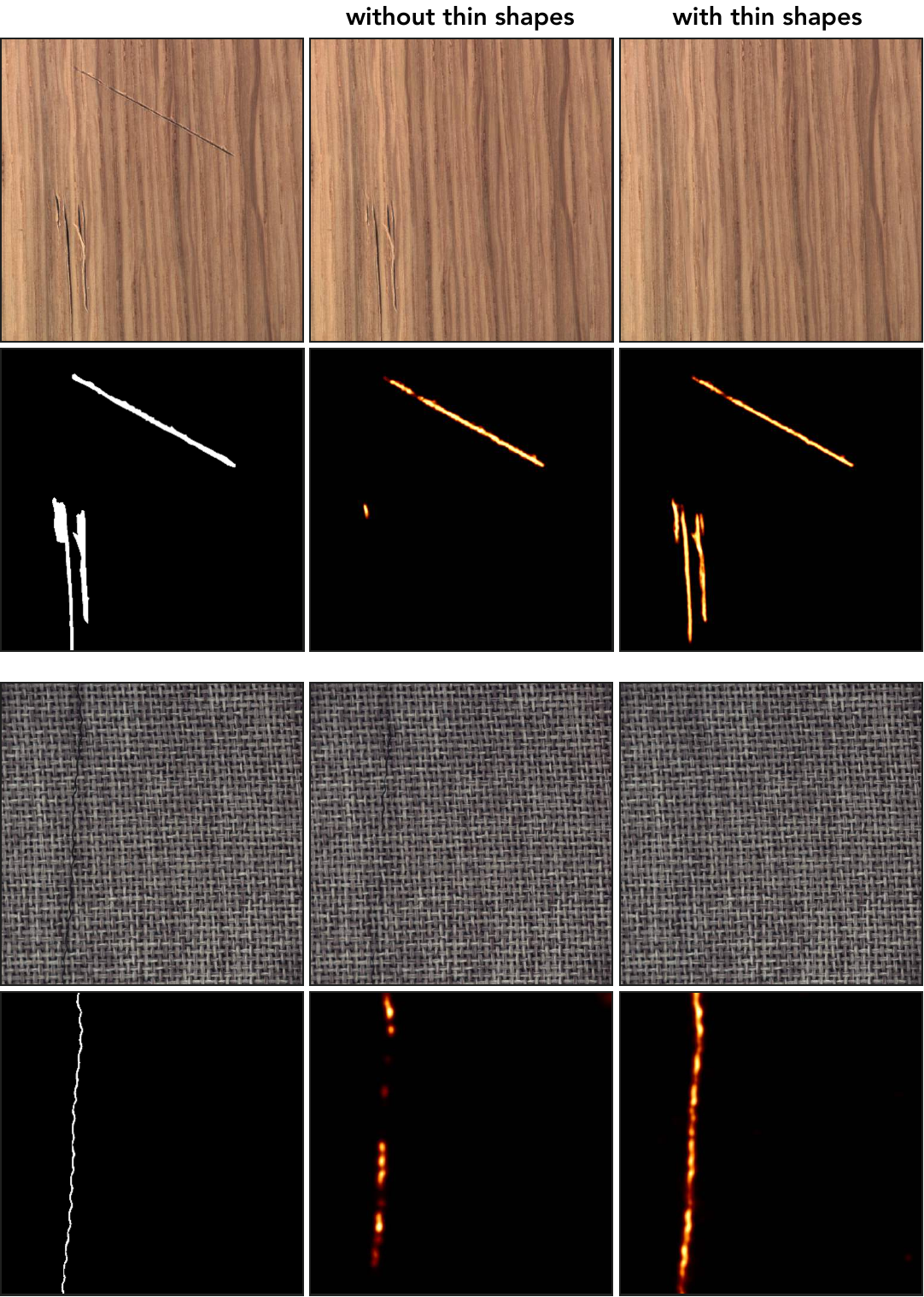}
\caption{
Illustration of the qualitative difference in filtering anomalous patterns during reconstruction 
for models trained with and without thin shapes. 
The first row (for each example) shows a defective input image and the corresponding reconstructions 
from models trained without and with thin shapes, respectively. 
The second row (for each example) shows the human annotation mask alongside the anomaly heatmaps 
produced by the two models.
}
\label{fig_bauer}
\end{figure}

As discussed in Section \ref{sec:section5}, incorporating thin, curve-like masks alongside broad occlusions
during training significantly improves the correction of the corresponding thin, curve-shaped anomalies
in the testing phase. Figure \ref{fig_bauer} illustrates two examples. Notably, models trained without
curve-shaped occlusions consistently fail to completely eliminate such anomalies in the affected regions.

\section{Ablation Study for Gaussian Noise Regularization}
We conduct an ablation study by evaluating the performance of models trained with and without Gaussian noise regularization. 
The corresponding results are reported in Table \ref{table_ablation}. 
Two key observations emerge. 
First, regularization with Gaussian noise consistently improves segmentation accuracy. 
Second, models trained without this regularization exhibit higher performance variance. 
While the relative improvement is moderate, ranging between $0.01$ and $1.45$, this is mainly because the corruption model alone is already responsible for near-saturated results.
Nevertheless, noise regularization remains important: it improves segmentation accuracy, stabilizes training by reducing the variance of the final results, and its effect is clearly visible in the enhanced reconstruction quality of the model. 

\begin{figure*}[t]
\centering
\includegraphics[scale = 0.78]{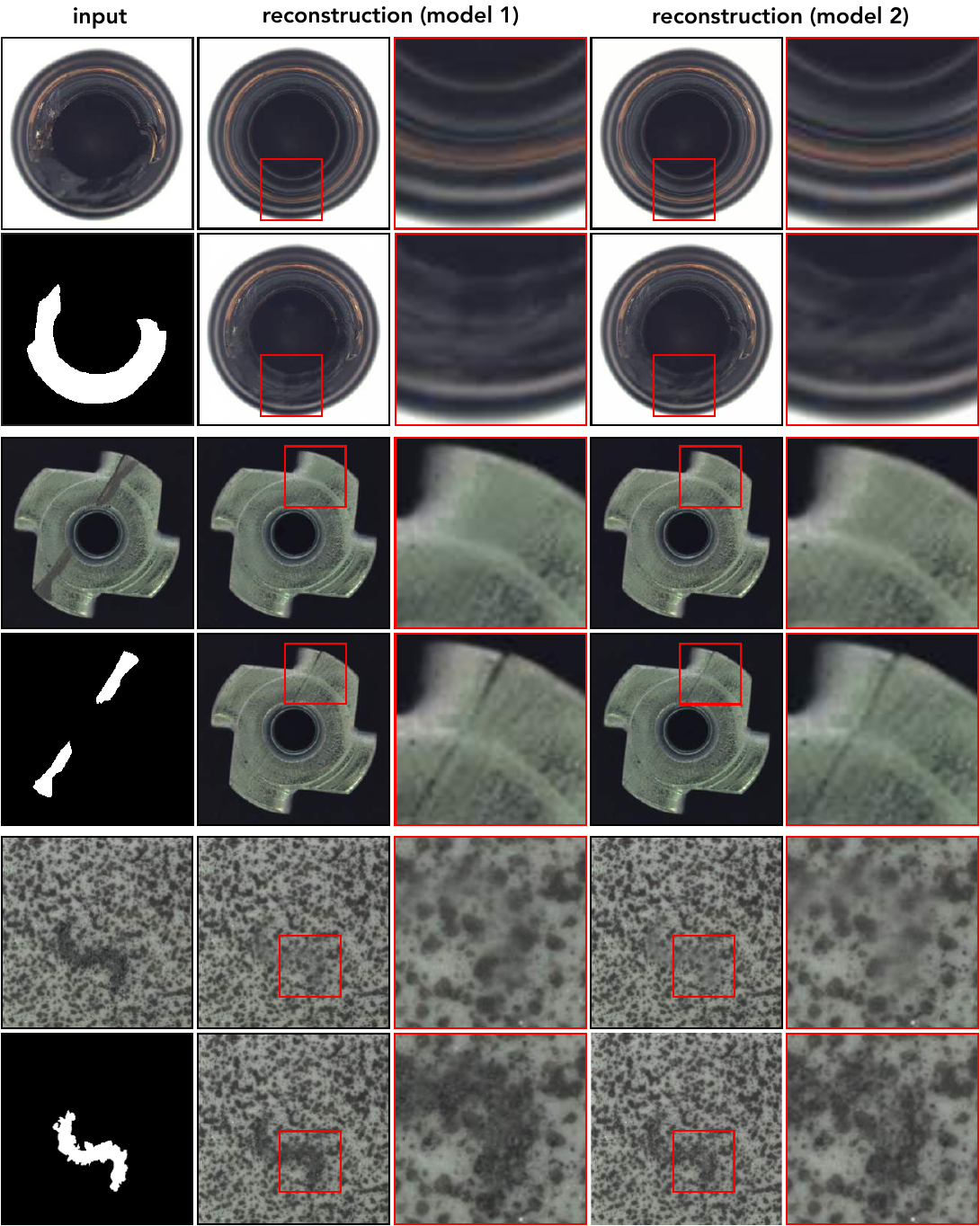}
\caption{
Qualitative comparison of reconstructions from models trained with and without Gaussian noise regularization. 
Odd rows show a defected input image followed by reconstructions from two regularized models, 
while even rows show the corresponding human-annotated anomaly mask followed by reconstructions from two unregularized models. 
For each reconstruction, a zoom-in of the region marked by the red rectangle is provided to enable detailed comparison. 
Gaussian noise regularization consistently enhances the correction of anomalous regions.
}
\label{fig_bauer5}
\end{figure*}

\begin{figure*}[t]
\centering
\includegraphics[scale = 0.87]{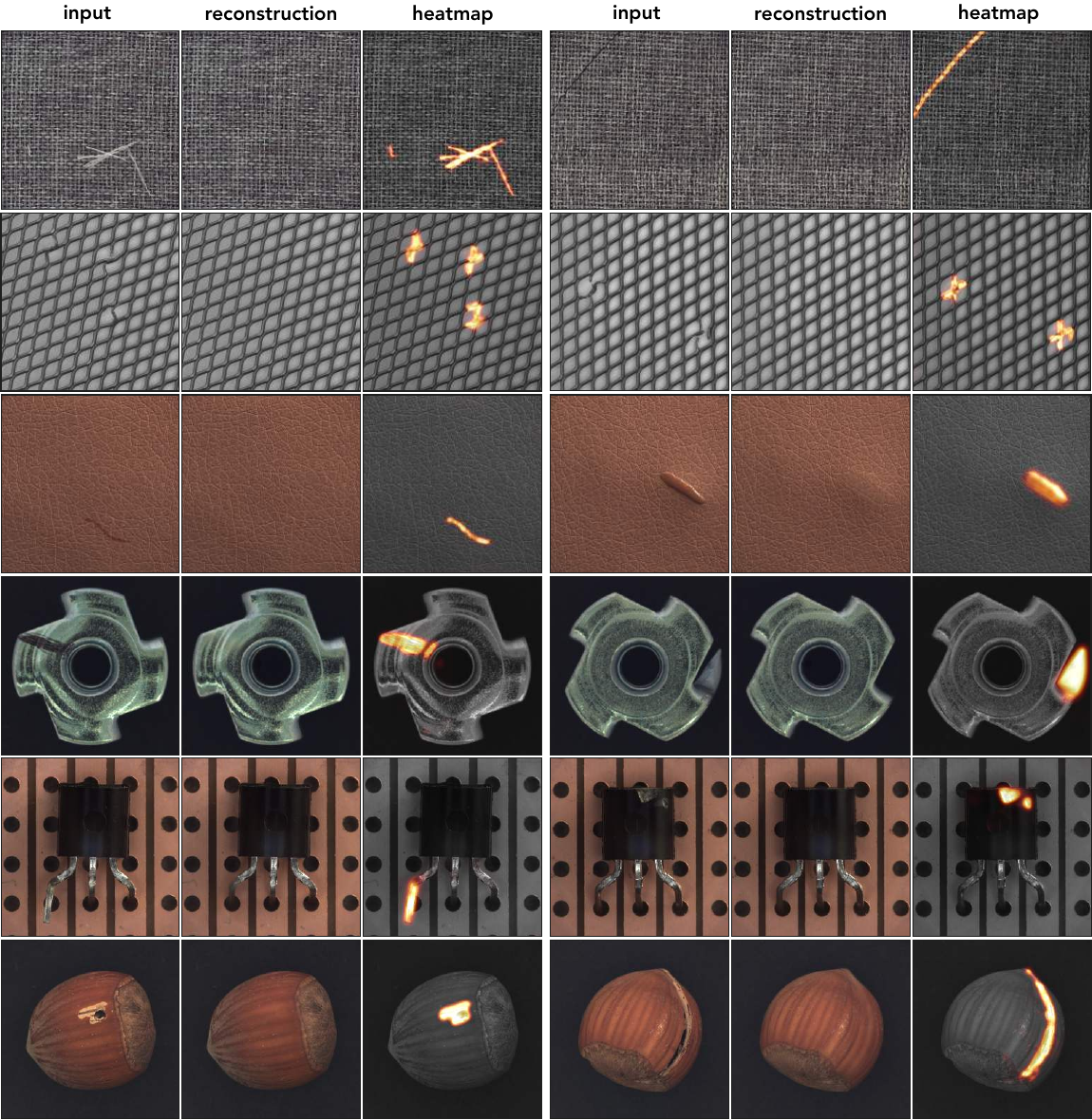}
\caption{
Qualitative results of reconstructions and corresponding anomaly heatmaps produced by our model. 
Each row shows two examples from one category. For each example we show the defective input image, 
its reconstruction, and the heatmap overlay.
}
\label{fig_examples}
\end{figure*}

\begin{figure*}[t]
\centering
\includegraphics[scale = 0.87]{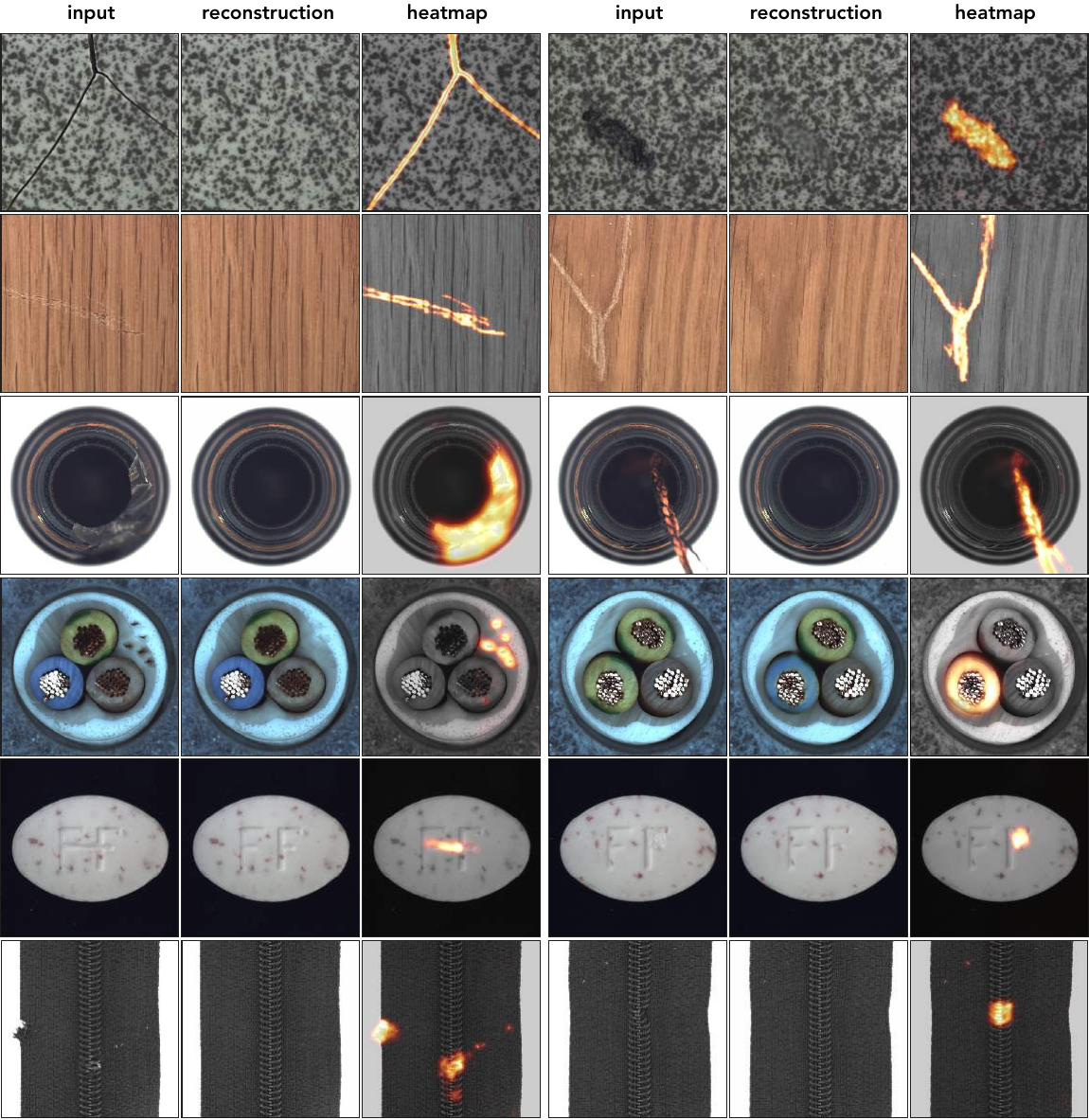}
\caption{
Qualitative results of reconstructions and corresponding anomaly heatmaps produced by our model. 
Each row shows two examples from one category. For each example we show the defective input image, 
its reconstruction, and the heatmap overlay.
}
\label{fig_examples2}
\end{figure*}

In anomaly detection, perfect reconstructions are not always required, since even small pixel-level differences between input and output can suffice to localize anomalies. 
However, the reconstructions obtained with noise regularization are noticeably more faithful and less artifact-prone, underscoring its relevance even when the quantitative gains appear modest. 
Figure \ref{fig_bauer5} illustrates this qualitative improvement, while Figure \ref{fig_examples} and \ref{fig_examples2} present further examples and highlights the quality of the resulting heatmaps.

\begin{table}[b]
  \caption{Ablation study with and without Gaussian noise regularization 
  measured with \textbf{P-AUROC} on MVTec AD.}
  \label{table_ablation}
  \centering
  \scalebox{0.9}{
  \begin{tabular}{lccc}
    \toprule
    Category & With noise & W/o noise & Improvement \\
    \midrule
    carpet      & 99.73 $\pm$ 0.02 & 99.20 $\pm$ 0.15 & +0.53 \\
    grid          & 99.79 $\pm$ 0.01 & 99.75 $\pm$ 0.06 & +0.04 \\
    leather     & 99.75 $\pm$ 0.00 & 99.62 $\pm$ 0.08 & +0.13 \\
    tile            & 98.75 $\pm$ 0.38 & 98.35 $\pm$ 0.65 & +0.40    \\
    wood        & 98.55 $\pm$ 0.03 & 98.43 $\pm$ 0.16 & +0.12 \\
    bottle        & 99.24 $\pm$ 0.05 & 98.60 $\pm$ 0.24 & +0.64 \\
    cable        & 98.62 $\pm$ 0.03 & 98.07 $\pm$ 0.33 & +0.55 \\
    capsule    & 99.18 $\pm$ 0.08 & 98.09 $\pm$ 0.11 & +0.09    \\
    hazelnut   & 99.48 $\pm$ 0.07 & 99.03 $\pm$ 0.15 & +0.45 \\
    metal nut  & 99.00 $\pm$ 0.56 & 97.55 $\pm$ 1.13 & +1.45 \\
    pill            & 99.52 $\pm$ 0.01 & 99.16 $\pm$ 0.09 & +0.36    \\
    screw       & 99.38 $\pm$ 0.28 & 99.33 $\pm$ 0.33 & +0.05    \\
    toothbrush & 99.45 $\pm$ 0.02 & 99.44 $\pm$ 0.03 & +0.01  \\
    transistor  & 98.12 $\pm$ 0.78 & 97.88 $\pm$ 1.02 & +0.24    \\
    zipper      & 99.70 $\pm$ 0.00 & 99.69 $\pm$ 0.00 & +0.01 \\
    \bottomrule
  \end{tabular}}
\end{table}

\end{document}